\documentclass[a4paper,conference]{IEEEtran}
\ifCLASSINFOpdf
\else
\fi
\usepackage{times}
\usepackage{epsfig}
\usepackage{graphicx}
\usepackage{amsmath}
\usepackage{amssymb}

\def\etal{\emph{et al.}}

\def\ie{\emph{i.e}}
\def\m{\mathcal}
\usepackage{subfig}
\usepackage{tablefootnote}
\usepackage{multirow}
\usepackage{colortbl}
\usepackage{capt-of}
\usepackage{fixltx2e}
\usepackage{floatrow}
\usepackage{enumitem}
\usepackage{multirow}
\usepackage{makecell}

\begin{document}
%
\title{Unsupervised Restoration of Weather-affected Images using Deep Gaussian Process-based CycleGAN}

	\author{Rajeev Yasarla\thanks{equal contribution} \qquad Vishwanath A. Sindagi\qquad Vishal M. Patel\\
		Johns Hopkins University\\
		Department of Electrical and Computer Engineering, Baltimore, MD 21218, USA\\
		{\tt\small \{ryasarl1, vishwanathsindagi, vpatel36\}@jhu.edu}
	}


%


\maketitle

\begin{abstract}
	Existing approaches for restoring weather-degraded images follow a fully-supervised paradigm and they require paired data for training. However, collecting paired data for weather degradations is extremely challenging, and existing methods end up training on synthetic  data. To overcome this issue, we describe an approach for supervising deep networks that is based on CycleGAN, thereby enabling the use of unlabeled real-world data for training. Specifically, we introduce new losses for training CycleGAN that lead to more effective training, resulting in high quality  reconstructions. These new losses are obtained by jointly modeling the latent space embeddings  of  predicted clean images and original clean images through Deep Gaussian Processes. This enables the CycleGAN architecture to transfer the knowledge from one domain (weather-degraded) to another (clean) more effectively. We demonstrate that the proposed method  can be effectively applied to different restoration tasks like de-raining, de-hazing and de-snowing and it outperforms other unsupervised techniques (that leverage weather-based characteristics) by a considerable margin.
\end{abstract}

\section{Introduction}

\label{sec:intro}
Weather conditions such as rain, fog (haze) and snow are aberrations in the environment that adversely affect the light rays traveling from the object to a visual sensor \cite{Authors16,Authors17f,he2010single,dehaze_2008,Authors17b,nagarajan2018simple}. This typically causes   detrimental effects on the images captured by the sensors, resulting in poor aesthetic quality. Additionally, such images also reduce the performance of down-stream computer vision tasks such as detection and recognition \cite{Li2018BenchmarkingSD}. Such tasks are  often critical parts in autonomous navigation systems, which emphasizes the need to  address these degradations. These reasons has motivated a plethora of research on methods to remove such effects. 

\begin{figure}[htp!]
	\begin{flushleft}
		\includegraphics[width=\linewidth]{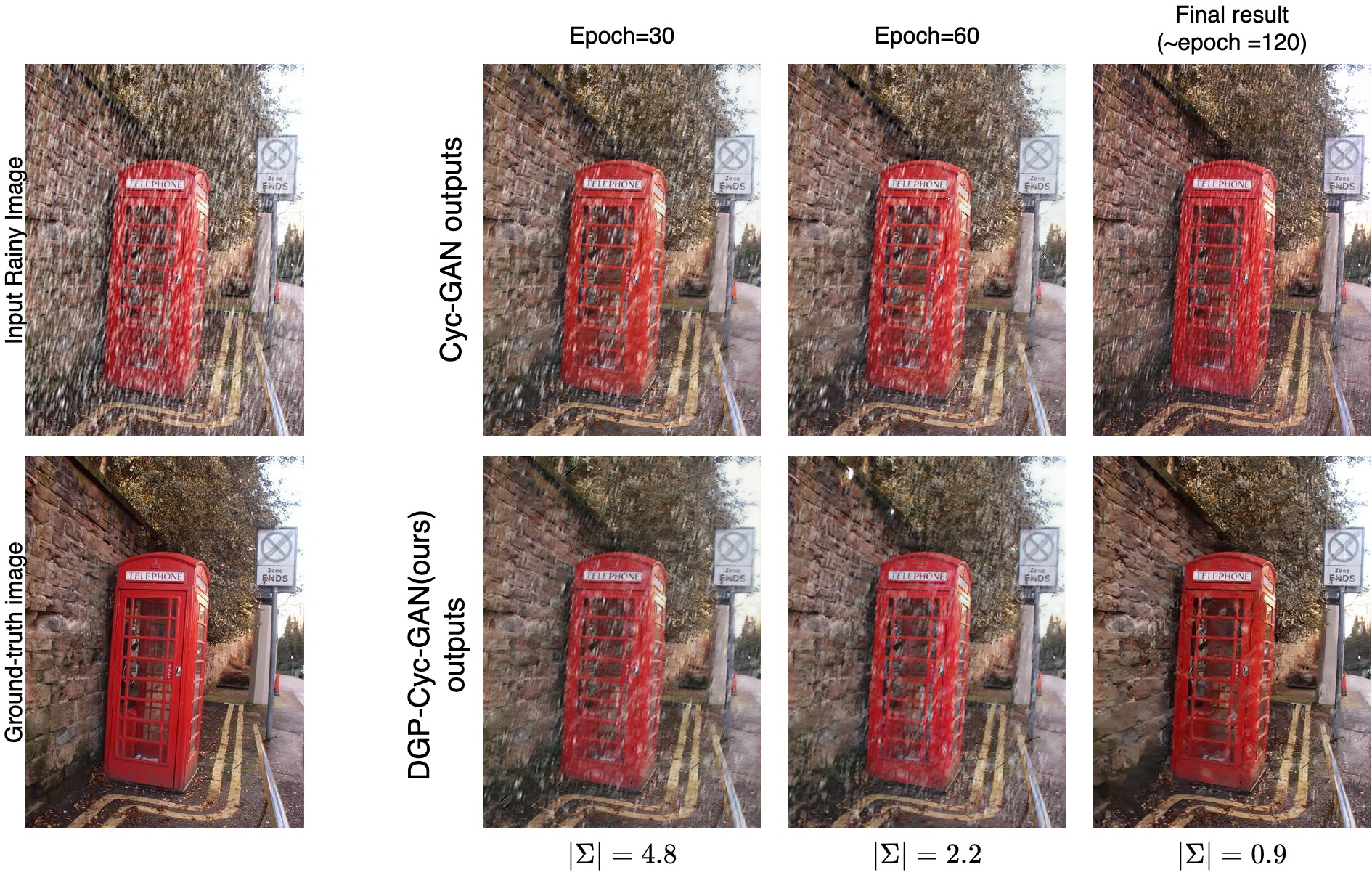}\\
	\end{flushleft}
	\vskip -20pt 
	\caption{Results ($\tilde{I}_c$) of the forward mapping ($\m{F}_{W\rightarrow C}$) network on a sample rain image for the deraining task from Cyc-GAN (\textit{top-row}) and DGP-Cyc-GAN (\textit{bottom-row}) at different epochs. These images are used as input to the reverse mapping network $\m{F}_{C\rightarrow W}$. Note that these images are noisy in the initial epochs which leads to incorrect training. Using Deep-GP, we are able to estimate the uncertainty which is in-turn incorporated into the loss function through a multiplier, resulting in the use of only non-noisy samples for the supervision. Hence, the proposed method outputs progressively cleaner outputs with lower uncertainty.} 
	\label{fig:sample_ic_epochs}
\end{figure}

Recent research on weather-based restoration (de-raining, de-hazing and de-snowing)  is typically focused on designing   convolutional neural network (CNN)-based pixel-to-pixel regression architectures. These works typically   incorporate different aspects such as attention \cite{Authors18b,liu2019griddehazenet}, degradation characteristics \cite{yang2018proximal,dehaze_iccv17}, and many more.  While these approaches have been effective in achieving   high-quality restorations, they essentially follow a fully-supervised paradigm. Hence, they require paired  data to successfully train their networks. Considering that weather effects are naturally occurring phenomena, it is practically infeasible to collect data containing pairs of clean and weather-affected (rainy, hazy, snowy) images. Due to this issue, existing restoration networks are unable to  leverage real-world data, and they end up training on synthetically generated paired data.  However, networks trained on synthetic datasets suffer distributional shift due to which they perform poorly on   real-world images.

In this work, we address this issue by describing an approach to train a network from  a set of degraded and clean images which are not paired.  We view the restoration problem as a task of translation from one domain (weather-affected) to another (clean) and  build on a recent popular method - CycleGAN \cite{zhu2017unpaired}.  One may train a network with unpaired data by  ensuring  that the restored images belong to the same distribution as that of real clean images \cite{goodfellow2014generative}. However, the problem of translating images from one domain to another using unpaired data is ill-posed, as many mappings may satisfy this constraint.    CycleGAN addresses this issue by using a forward mapping (clean to weather-affected) and a reverse mapping network (weather-affected to clean), and introduces an additional constraints via cycle-consistency losses along with adversarial losses .  It is important to note that for  computing the consistency losses, the output of the forward mapping function is used as input to the reverse mapping function and vice versa. However, during the initial stages of the training process, these outputs are potentially noisy which can be detrimental to learning an accurate function to map a degraded image to clean image (see Fig.~\ref{fig:sample_ic_epochs} for details).

To overcome this issue, we introduce a new set of losses that provide additional supervision for predicted clean images in the latent space. These losses are derived by jointly modeling the latent projections of clean images and predicted clean images using Deep Gaussian Processes (GP). By conditioning this joint distribution on the projections of clean images, we are able to obtain pseudo-labels for the predicted clean images in the latent space. These pseudo-labels are then used in conjunction with the uncertainty derived from Deep-GP, to supervise the restoration network in the latent space. The use of uncertainty information ensures that only confident pseudo-labels are used during the training process. In other words, the noisy labels are disregarded, thus avoiding their use in the training of the forward and the reverse mapping networks in CycleGAN. To demonstrate the effectiveness of the proposed method, we conducted extensive experiments on multiple weather related restoration tasks like de-raining, dehazing and de-snowing using several benchmark datasets.


\noindent Following are the main contributions of our work:
\begin{itemize}[topsep=0pt,noitemsep]
	\item We address the problem of learning to restore images  for different weather conditions from unpaired data. Specifically, we present Deep Gaussian Processes-based CycleGAN that introduces new losses for training the  CycleGAN architecture  resulting in high quality restorations.  
	\item The proposed method achieves  significant improvements over CycleGAN and other unsupervised techniques for  various restoration tasks like as de-raining, de-hazing and de-snowing. 
	\item We show that the proposed method is able to leverage real-world data better as compared to CycleGAN by performing evaluation on a  down-stream task, namely   object detection on the restored images. 
	
\end{itemize}

\section{Related Work}

\noindent \textbf{Image restoration for weather-degradations}: Restoring weather-degraded images is a challenging problem since it is an  ill-posed task even when paired data is available. Typically, these degradations are modeled based on the principles of physics, and the solutions are obtained using    these physics-based models. Due to differences in the weather characteristics and the models, most existing   approaches  address these conditions separately. For example: 
\begin{enumerate}[topsep=0pt,noitemsep]
	\item Rain: A rainy image follows an additive model, where it is expressed as a superposition of a clean image and rain streaks \cite{Authors16,Authors17f,Authors18b,Authors18,yang2017deep,yasarla2019uncertainty,yasarla2020confidence,yasarla2020exploring,yasarla2021semi,valanarasu2021transweather,yasarla2022art}.  Existing approaches for de-raining   incorporate various aspects into their network design such as  rain characteristics \cite{Authors18}, attention \cite{Authors18b,wang2019spatial}, context-awareness \cite{Authors18d},  depth information \cite{hu2019depth}, and  semi-supervised learning\cite{yasarla2020syn2real,wei2019semi}. A comprehensive analysis of these methods can be found in \cite{li2019singlecomprehensive, Rain_PAMI_survey}.  
	\item Haze: A hazy image is modeled as a superposition of a  transmission map and an attenuated clean image \cite{he2010single,dehaze_2008,dehaze_iccv17,Li2018BenchmarkingSD}. Like de-raining techniques, approaches developed for image de-hazing exploit different concepts such as multi-scale fusion \cite{dehaze_mul_fusion,ren2016single1,dehaze_2016_eccv,ren2019single}, gated fusion \cite{ren2018gated}, network design \cite{zhang2018densely}, prior-information \cite{yang2018proximal,dehaze_iccv17,galdran2018duality}, adversarial loss \cite{li2018single,yang2018towards}, image-to-image translation \cite{qu2019enhanced}, and attention-awareness \cite{liu2019griddehazenet}. For  more details, the readers are referred to  \cite{dehaze_survey,li2017haze}. 
	\item Snow: A Snowy image is modeled similar to that of a rainy image, however the characteristics of snow-residue are quite distinct from rain-residue \cite{Authors17b,nagarajan2018simple}. Hence,  approaches like \cite{liu2018desnownet,li2019single,huang2020single} exploit various properties of snow to perform high-quality de-snowing.  
\end{enumerate}

While these approaches are able to achieve superior restoration quality, they essentially follow a  fully-supervised paradigm and cannot be used for training on unpaired data. Additionally, these techniques are weather-specific as they incorporate weather-related models pertaining to a particular weather condition. In contrast, we propose a more general network that can be trained on unpaired data for any weather condition.

\noindent \textbf{Unpaired image-to-image translation}: Initial approaches for image-to-image translation \cite{isola2017image,zhu2017toward} are   based on generative adversarial networks \cite{goodfellow2014generative}. However, these methods employ paired data to train their networks. To overcome this issue, several techniques \cite{zhu2017unpaired,yi2017dualgan,liu2017unsupervised,kim2017learning} have been proposed   for training networks with unpaired data. Zhu \etal \cite{zhu2017unpaired} proposed CycleGAN which introduced cycle-consistency loss to impose an  additional constraint that each image should be reconstructed correctly when translated twice. The objective is to  conserve the overall structure and content of the image.   DualGAN \cite{yi2017dualgan} and DiscoGAN \cite{kim2017learning} follow a similar approach, with slightly different losses. In contrast, approaches like \cite{liu2016coupled,liu2017unsupervised} consider a shared-latent space and they learn a joint distribution over
images from two domains.  They assume that images from two domains can be mapped into a low-dimensional shared-latent space. Most of the subsequent works \cite{gonzalez2018image,kondo2019flow,mejjati2018unsupervised,kazemi2018unsupervised,amodio2019travelgan} build on these approaches by incorporating additional information or structure like domain/feature disentanglement \cite{gonzalez2018image,kondo2019flow}, attention-awareness \cite{mejjati2018unsupervised}, learning of domain invariant representation \cite{kazemi2018unsupervised} and instance-awareness \cite{amodio2019travelgan}.

\noindent \textbf{Unpaired restoration for weather-degradations}: Compared to fully-supervised  approaches, research on unpaired restoration (for weather-degradations) has received limited attention. Most of the existing efforts are inspired by the unpaired translation approaches like CycleGAN \cite{zhu2017unpaired} and DualGAN \cite{yi2017dualgan}. These approaches typically exploit weather-specific characteristics, and hence are designed individually for different weather conditions.  For example, de-raining approaches like  \cite{han2020decomposed,jin2019unsupervised,wei2019deraincyclegan,zhu2019singe} use rain properties to  decompose the de-raining problem into foreground/background separation and employ rain-mask to provide additional supervision to train the CycleGAN network. Similarly, de-hazing approaches like \cite{engin2018cycle,dudhane2019cdnet,yang2018towards} extend CycleGAN  by incorporating haze related features. For example, Yang \etal \cite{yang2018towards} and Dudhane \etal \cite{dudhane2019cdnet} employ physics-based haze model to improve disentanglement and reconstruction in CycleGAN. Since these  are designed specifically for a particular weather condition, they do not generalize to other conditions. In contrast, we propose a more general method that does not assume any specific weather-related model or characteristics.  We enforce additional supervision during the training of CycleGAN which enables us to learn more accurate restoration functions resulting in better  performance.


\vspace{-0.5em}
\section{Proposed method}

\noindent \textbf{Preliminaries}: We are given a dataset of unpaired data, $\m{D}=\m{D}_w \cup \m{D}_c$, where $\m{D}_w=\{I_w^i\}_{i=1}^N$ consists of a set of images degraded due to a particular weather condition and $\m{D}_c=\{I_c^i\}_{i=1}^N$ consists of a set of clean images. The goal is to learn a restoration function $\m{F}_{W\rightarrow C}$, that maps a weather-degraded image ($I_w$) to a clean image ($I_c$). Since  CycleGAN \cite{zhu2017unpaired} enables training from unpaired data, we use this approach as a starting point to learn this function. In the CycleGAN framework, restoration function $\m{F}_{W\rightarrow C}$ corresponds to the forward mapping network. As discussed earlier,  CycleGAN  enforces two constraints:  (i) the distribution of restored images $P(\tilde{I_c})$ is similar to that of clean images $P(I_c)$ and this is achieved with the aid of adversarial loss \cite{goodfellow2014generative}, (ii) In addition to $\m{F}_{W\rightarrow C}$, it also learns a  reverse  mapping function  $\m{F}_{C\rightarrow W}$ and ensures cycle consistency which is defined as:  $\m{F}_{C\rightarrow W}(\m{F}_{W\rightarrow C}(I_w))=I_w$. 

In order to achieve high restoration quality, we need to learn an accurate restoration function ($\m{F}_{W\rightarrow C}$). Although CycleGAN enforces the aforementioned constraints, these are not necessarily sufficient.  This is because, in the case of CycleGAN, the results ($\tilde{I}_c$) of the forward mapping ($\m{F}_{W\rightarrow C}$) network  are used as input to the reverse mapping network $\m{F}_{C\rightarrow W}$. These images are typically noisy in the initial epochs, as shown in Fig.~\ref{fig:sample_ic_epochs} which leads to incorrect supervision and the network will overfit to the noisy data. In this work, we attempt to provide additional supervision via a set of new losses to overcome the aforementioned issues, thereby resulting in better restoration quality.

Further, although  our method is based on CycleGAN, we demonstrate that it can be applied to other unpaired translation approaches like UNIT GAN (see supplementary).  In what follows, we describe the proposed approach in detail.

\begin{figure*}[htp!]
	\begin{center}
		\includegraphics[width=.88\linewidth]{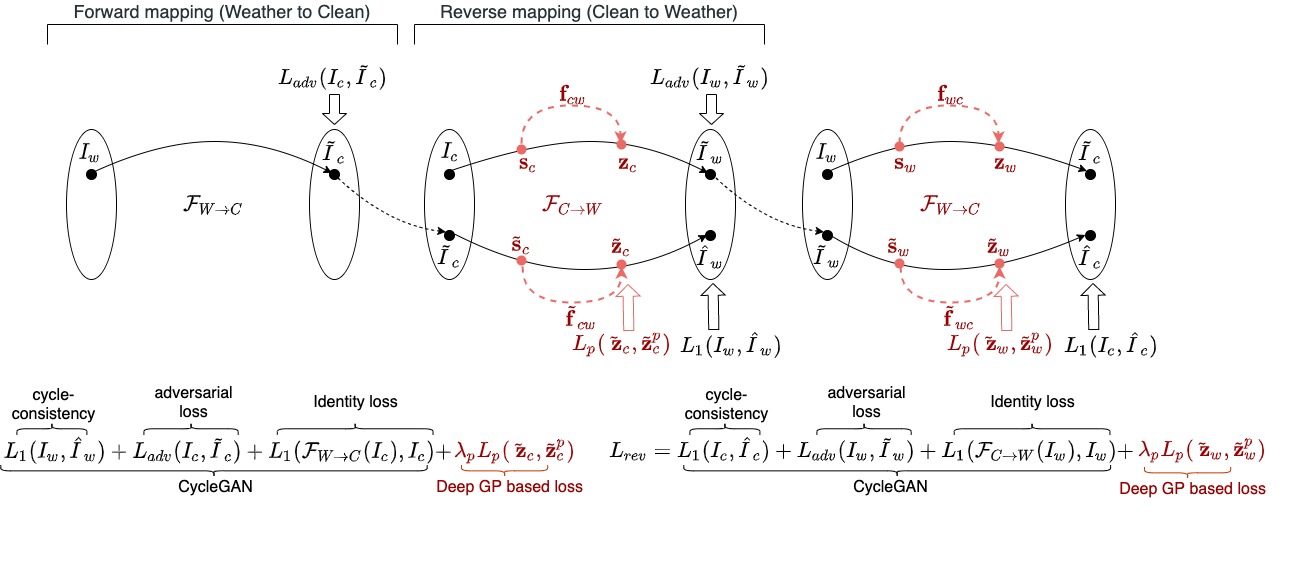}
	\end{center}
	\vskip -35pt 
	\caption{Overview of the proposed Deep Gaussian Processes-based CycleGAN.}
	\label{fig:overview}
\end{figure*}

\noindent \textbf{Deep Gaussian Process-based CycleGAN}: Fig.~\ref{fig:overview} shows an overview of the proposed method. As it can be observed, we build on CycleGAN. We introduce additional losses in the latent space for the forward and reverse mapping (as shown in red color).   We extract latent space embeddings from two intermediate layers  ($\mathbf{s}$ and $\mathbf{z}$) in both the networks (forward mapping and reverse mapping). That is, in the forward mapping network ($\m{F}_{W\rightarrow C}$), a weather-degraded image $I_w$ is mapped to a latent embedding $\mathbf{s_w}$, which is then mapped  to another embedding $\mathbf{z_w}$, before being mapped  to restored (clean) image $\tilde{I}_c$. The restored (cleaned) image $\tilde{I}_c$ is then forwarded through the reverse mapping network ($\m{F}_{C\rightarrow W}$) to produce latent vectors $\mathbf{\tilde{s}_c}$ and $\mathbf{\tilde{z}_c}$, before being mapped to a reconstructed weather-degraded image $\hat{I}_w$.   Similarly, a clean image  $I_c$ is mapped to a latent embedding $\mathbf{s_c}$, which is then mapped  to another embedding $\mathbf{z_c}$, before being mapped  to  reconstructed weather-degraded image $\tilde{I}_w$. 

For learning to reconstruct $\tilde{I}_c$, CycleGAN provides supervision by enforcing cycle consistency ($L_{fwd}^{cyc}=|I_w - \hat{I}_w|_1$) and adversarial loss ($L_{fwd}^{adv}(I_c, \tilde{I}_c)$). In order to ensure appropriate training, we provide additional supervision in the latent space  by deriving pseudo-labels  for the latent projection $\mathbf{\tilde{z}}_c$ of $\tilde{I}_c$. The  pseudo-label $\mathbf{\tilde{z}}^{p}_c$ is obtained by expressing the latent vectors of restored clean images $\tilde{I}_c$ in terms of projections of the original clean images $I_c$ by modeling joint distribution using Gaussian Process (GP). That is, given a set of ``clean image'' latent vectors $\mathbf{s}_c$, from the first intermediate layer in the deep network, we write $\textbf{z}_c$ (latent vector of ``clean images" from second intermediate layer) as a function of $\mathbf{s}_c$ as $\mathbf{z}_{c}=\mathbf{f}_{cw}\left(\mathbf{s}_{c}\right)$. Hence, for the ``restored clean image'' latent vectors $\mathbf{\tilde{s}}_c$, we can obtain the corresponding $\mathbf{\tilde{z}}_c^p$ using: $\mathbf{\tilde{z}}^{p}_{c}=\mathbf{\tilde{f}}_{cw}\left(\mathbf{\tilde{s}}_{c}\right).$


We aim to learn the  function $\mathbf{\tilde{f}}_{cw}$ via Gaussian Processes. 
Specifically,  we formulate the joint distribution of $\mathbf{{f}}_{cw}$ and $\mathbf{\tilde{f}}_{cw}$ (or correspondingly $\mathbf{{z}}^{}_c$ and $\mathbf{\tilde{z}}^{p}_c$)  as follows:

%
\setlength{\belowdisplayskip}{0pt} \setlength{\belowdisplayshortskip}{0pt}
\setlength{\abovedisplayskip}{0pt} \setlength{\abovedisplayshortskip}{0pt}
\begin{equation}
\resizebox{1.0\hsize}{!}{%
	$\left[\begin{array}{l}
	\mathbf{f}_{cw} \\
	\mathbf{\tilde{f}}_{cw}
	\end{array}\right]=
	\left[\begin{array}{l}
	\mathbf{z}_{c}\\
	\mathbf{\tilde{z}}^{p}_{c}
	\end{array}\right]=\mathcal{N}\left(\left[\begin{array}{l}
	\boldsymbol{\mu}_{c} \\
	\boldsymbol{\tilde{\mu}}_{{c}}
	\end{array}\right],\left[\begin{array}{cc}
	K\left(\mathbf{s}_{c}, \mathbf{s}_{c}\right) & K\left(\mathbf{s}_{c}, \mathbf{\tilde{s}}_{c}\right) \\
	K\left(\mathbf{\tilde{s}}_{c},\mathbf{s}_{c}\right) & K\left(\mathbf{\tilde{s}}_{c},\mathbf{\tilde{s}}_{c}\right)
	\end{array}\right]+\sigma_{\epsilon}^{2}\mathbb{I}\right).
	$}
\end{equation}
Here, $K$ is the kernel matrix $K(\mathbf{U},\mathbf{V})_{i,j} = k(\mathbf{u}_i,\mathbf{v}_j)$, where $\mathbf{u}_i$ is the $i^{th}$  vector of $\mathbf{U}$ and $\mathbf{v}_j$ is the $j^{th}$  vector of $\mathbf{V}$, and $\mathbb{I}$ is identity matrix, $\sigma_{\epsilon}^{2}$ is  the additive noise variance that is set to 0.01.
By conditioning the above distribution, we obtain the following distribution for $\mathbf{z}^{p}_c$:

\begin{equation}
P\left(\mathbf{\tilde{z}}^{p}_{c} | \mathbf{X}^{z}_{c},\mathbf{X}^{s}_{c}, \mathbf{\tilde{s}}_{c}\right)=\mathcal{N}\left(\boldsymbol{\tilde{\mu}}_{c},\mathbf{\tilde{\Sigma}}_{{c}}\right),
\end{equation}

\noindent where, $\mathbf{X}^{s}_{c}$ and $\mathbf{X}^{z}_{c}$ are matrices of latent projections $\mathbf{s}_c$'s and $\mathbf{z}_c$'s, respectively of all clean images in the dataset, and $\boldsymbol{\tilde{\mu}}_{c},\mathbf{\tilde{\Sigma}}_{{c}}$ are defined as follows:

\begin{equation}
\label{eq:muc_orig}
\resizebox{1.0\hsize}{!}{
	$
	\begin{aligned}
	\boldsymbol{\tilde{\mu}}_{{{c}}}&=K\left(\mathbf{\tilde{s}}_{c}, \mathbf{X}^{s}_{c}\right)\left[K\left(\mathbf{X}^{s}_{c}, \mathbf{X}^{s}_{c}\right)+\sigma_{\epsilon}^{2} \mathbb{I}\right]^{-1} \mathbf{X}^{z}_{c}, \\
	\mathbf{\tilde{\Sigma}}_{{{c}}}&=K\left(\mathbf{\tilde{s}}_{c}, \mathbf{\tilde{s}}_{c}\right)-K\left(\mathbf{\tilde{s}}_{c}, \mathbf{X}^{s}_{c}\right)\left[K\left(\mathbf{X}^{s}_{c}, \mathbf{X}^{s}_{c}\right)+\sigma_{\epsilon}^{2} \mathbb{I}\right] 
	K\left(\mathbf{X}_{s_{c}}, \mathbf{\tilde{s}}_{c}\right) +\sigma_{\epsilon}^{2}\mathbb{I}.\\
	\end{aligned}
	$}
\end{equation}

\noindent We use the squared exponential kernel:  
\begin{equation}
\label{eq:sqe}
k\left(\mathbf{x}_{i}, \mathbf{x}_{j}\right)=\beta^{2} \exp \left(-\frac{\left\|x_{i}-x_{j}\right\|^{2}}{2 \gamma^{2}}\right),
\end{equation}
where $\beta$ is the signal magnitude and $\gamma$ is the length scale. In  all our experiments, we use $\frac{\beta}{\gamma}=1.0$. 

Considering that Deep Gaussian Processes have better representation power \cite{damianou2013deep} as compared to single layer Gaussian Processes, we model  $\left[\begin{array}{c}
\mathbf{f}_{c w} \\
\tilde{\mathbf{f}}_{c w}
\end{array}\right]$ using Deep GP with $L$ layers as follows:   

\begin{equation}
\label{eq:dgp_original}
\begin{aligned}
\mathbf{\bar{f}}_{cw} = 
\left[\begin{array}{c}
\mathbf{f}_{c w} \\
\tilde{\mathbf{f}}_{c w}
\end{array}\right]
&\sim G P\left(\boldsymbol{\mu}^{L}_f, K^L\left(\mathbf{h}^{L-1},\mathbf{h}^{L-1}\right)\right),\\
\mathbf{h}^l&\sim G P\left(\boldsymbol{\mu}^{l}_h, K^{l}\left(\mathbf{h}^{l-1},\mathbf{h}^{l-1}\right)\right),\\
\mathbf{h}^1&\sim G P\left(\mathbf{\mu}^{1}_h, K^1\left(\mathbf{s}_c,\mathbf{s}_c\right)\right),
\end{aligned}
\end{equation} 
where, $l$ indicates the layer index and $\mathbf{h}^l$ indicates the $l^{th}$ hidden layer. The use of Deep GPs along with convolutional neural networks  (CNNs)  is inspired by the works of \cite{van2017convolutional,bradshaw2017adversarial,tran2019calibrating}. The joint distribution can be written as:
\begin{equation}
\resizebox{0.9\hsize}{!}{$
\begin{aligned}
P\left(\mathbf{\bar{f}}, \mathbf{h}^{1: L} | \mathbf{X}^z_{c},\mathbf{X}^s_{c},\mathbf{\tilde{s}}_{c}\right)=P\left(\mathbf{f} | \mathbf{h}^{L}\right) &P\left(\mathbf{h}^{L} | \mathbf{h}^{L-1}\right) \ldots\\
&\ldots P\left(\mathbf{h}^{1} | \mathbf{X}^z_{c},\mathbf{X}^s_{c},\mathbf{\tilde{s}}_{c}\right)
\end{aligned}.$}
\end{equation}
Marginalizing the above distribution over $\mathbf{h}$, we obtain: 
\begin{equation}
P(\mathbf{f} |\mathbf{X}^z_{c},\mathbf{X}^s_{c},\mathbf{\tilde{s}}_{c})=\int P\left(\mathbf{f}, \mathbf{h}^{1: L} | \mathbf{X}^z_{c},\mathbf{X}^s_{c},\mathbf{\tilde{s}}_{c}\right) d \mathbf{h}^{1: L}.
\end{equation}
We approximate Deep GP with a single layer GP as described in \cite{lu2019interpretable}. More specifically, the authors in \cite{lu2019interpretable} proposed to  approximate Deep GP as a GP by calculating the exact moment. They provide general recipes for deriving the effective kernels for Deep GP of two, three, or infinitely many layers, composed of homogeneous or heterogeneous kernels. Their approach enables us to analytically integrate yielding effectively deep, single layer kernels. Based on this, we can rewrite the expression for $\mathbf{\bar{f}}$ (from Eq.~\ref{eq:dgp_original}) and consequently $\boldsymbol{\tilde{\mu}}_c$ and $\boldsymbol{\tilde{\Sigma}}_c$ using effective kernel (from Eq.~\ref{eq:muc_orig}) as follows:
\begin{equation}
\label{eq:final_mu}
\resizebox{1.0\hsize}{!}{
	$
	\begin{aligned}
	\mathbf{\bar{f}}&=\left[\begin{array}{c}
	\mathbf{f}_{c w} \\
	\tilde{\mathbf{f}}_{c w}
	\end{array}\right] 
	\sim G P\left(\left[\begin{array}{l}
	\boldsymbol{\mu}^L_{f} \\
	\boldsymbol{\tilde{\mu}}^L_{{f}}
	\end{array}\right], \left[\begin{array}{cc}
	K_{eff}\left(\mathbf{s}_{c}, \mathbf{s}_{c}\right) & K_{eff}\left(\mathbf{s}_{c}, \mathbf{\tilde{s}}_{c}\right) \\
	K_{eff}\left(\mathbf{\tilde{s}}_{c},\mathbf{s}_{c}\right) & K_{eff}\left(\mathbf{\tilde{s}}_{c},\mathbf{\tilde{s}}_{c}\right)
	\end{array}\right]+\sigma_{\epsilon}^{2}\mathbb{I}\right),\\
	\boldsymbol{\tilde{\mu}}_{{{c}}}&=K_{eff}\left(\mathbf{\tilde{s}}_{c}, \mathbf{X}^{s}_{c}\right)\left[K_{eff}\left(\mathbf{X}^{s}_{c}, \mathbf{X}^{s}_{c}\right)+\sigma_{\epsilon}^{2} \mathbb{I}\right]^{-1} \mathbf{X}^{z}_{c}, \\
	\mathbf{\tilde{\Sigma}}^{z}_{{c}}&=K_{eff}\left(\mathbf{\tilde{s}}_{c}, \mathbf{\tilde{s}}_{c}\right)-K_{eff}\left(\mathbf{\tilde{s}}_{c}, \mathbf{X}^{s}_{c}\right)\left[K_{eff}\left(\mathbf{X}^{s}_{c}, \mathbf{X}^{s}_{c}\right)+\sigma_{\epsilon}^{2} \mathbb{I}\right] 
	K_{eff}\left(\mathbf{X}^{s}_{c}, \mathbf{\tilde{s}}_{c}\right)+\sigma_{\epsilon}^{2}\mathbb{I}.
	\end{aligned}
	$}
\end{equation}
As described in \cite{lu2019interpretable},  the  effective kernel for a $L$-layer Deep GP  can be written as: $$k_{\mathrm{eff}}^{(L)}\left(\mathbf{x}_{i}, \mathbf{x}_{j}\right)=\frac{\beta_{L}^{2}}{\sqrt{1+2\left(\gamma_{l}^{-2}\right)\left[\beta_{L-1}^{2}-k_{\mathrm{eff}}^{(L-1)}\left(\mathbf{x}_{i}, \mathbf{x}_{j}\right)\right]}}.$$
Although we focus on the squared exponential kernel here, we experiment with other kernels  as well (see supplementary material). 

We  use the expression for $\boldsymbol{\tilde{\mu}}_c$ from Eq.~\ref{eq:final_mu} as  $\mathbf{\tilde{z}}^{p}_c$. We then use $\mathbf{\tilde{z}}^{p}_c$ to supervise  $\mathbf{\tilde{z}_c}$ and define the loss function as follows:
\begin{equation}
\label{eq:lpfwd}
{L}^{p}_{fwd} = \left(\mathbf{\tilde{z}}_{c}-\mathbf{\tilde{z}}^{p}_{c}\right)^{\textrm{T}}\left(\mathbf{\tilde{\Sigma}}^{z}_{{c}}\right)^{-1}\left(\mathbf{\tilde{z}}_{c}-\mathbf{\tilde{z}}^{p}_{c}\right) + \log\left(\det{|\mathbf{\tilde{\Sigma}}^{z}_{{c}}|}\right).
\end{equation}

To summarize, given the latent embedding vectors corresponding to  ``clean images'' from the first  and second intermediate layers ($\mathbf{s}_c$ and $\textbf{z}_c$, respectively), and latent embedding vectors corresponding to  ``restored clean images'' from the first intermediate layer $\mathbf{\tilde{s}}_c$, we obtain the pseudo-labels $\mathbf{\tilde{z}}_c^p$  using  Eq.~\ref{eq:final_mu}. These pseudo-labels are then used to supervise at the second intermediate layer in the network $\mathbf{z}_{c}$. Please see supplementary material for detailed algorithm.

Similarly, we can derive additional losses for the reverse mapping (see supplementary for details) as follows: 
\begin{equation}
\label{eq:lprev}
{L}^p_{rev} = \left(\mathbf{\tilde{z}}_{w}-\mathbf{\tilde{z}}^{p}_{w}\right)^{\textrm{T}}\left(\mathbf{\tilde{\Sigma}}^{z}_{{w}}\right)^{-1}\left(\mathbf{\tilde{z}}_{w}-\mathbf{\tilde{z}}^{p}_{w}\right) + \log\left(\det{|\mathbf{\tilde{\Sigma}}^{z}_{{w}}|}\right).
\end{equation}

\begin{table*}[t!]	
	\caption{Results for de-raining on real-world dataset (SPANet~\cite{wang2019spatial}). Higher numbers indicate better performance.}
	\label{tab:real_spanet}
	\vskip -10pt
	\huge
	\renewcommand{\arraystretch}{1.1} 
	\begin{center}
		\resizebox{1\textwidth}{!}{
			\begin{tabular}{ll|cccc|c|ccc|c}
				\hline
				\textbf{Type}  &  & \multicolumn{4}{c|}{Supervised} & Semi-supervised & \multicolumn{3}{c|}{Unsupervised} & \\ \cline{1-10} 
				\textbf{Dataset} & \textbf{Metric} & BaseNet& \textbf{SPANet}\cite{wang2019spatial} & \textbf{PreNet}\cite{ren2019progressive} & \textbf{MSPFN}\cite{Kui_2020_CVPR} & \textbf{SIRR}\cite{wei2019semi} & \textbf{Derain-CycleGAN}\cite{wei2019deraincyclegan} & \textbf{Cyc-GAN}\cite{zhu2017unpaired} & \textbf{Ours} & \textbf{Oracle}\\
				\hline
				SPANet~\cite{wang2019spatial} & PSNR/SSIM & \textbf{30.4/0.88}  & 33.6/0.92 &  33.2 / 0.91  & 33.8/0.93 & 33.3/0.93 & 34.1/ 0.95 & 32.4/0.86 & \textbf{36.4/0.95}   & 37.1/097 \\ \hline
			\end{tabular}
			
		}
	\end{center}
	\vspace{-0.5em}
\end{table*}

\begin{table*}[t!]	
	\caption{Results for de-raining on real-world dataset (SIRR~\cite{wei2019semi}). Lower numbers indicate better performance.}
	\label{tab:real}
	\vskip -12pt
	\huge
	\renewcommand{\arraystretch}{1.1} 
	\begin{center}
		\resizebox{.8\textwidth}{!}{
			\begin{tabular}{ll|cccc|c|cc}
				\hline
				\textbf{Type}  &  & \multicolumn{4}{c|}{Supervised} & Semi-supervised & \multicolumn{2}{c}{Unsupervised} \\ \cline{1-9} 
				\textbf{Dataset} & \textbf{Metric} & BaseNet & \textbf{SPANet}\cite{wang2019spatial} & \textbf{PreNet}\cite{ren2019progressive} & \textbf{MSPFN}\cite{Kui_2020_CVPR} & \textbf{SIRR}\cite{wei2019semi} &  \textbf{Cyc-GAN}\cite{zhu2017unpaired} & \textbf{Ours} \\
				\hline
				SIRR~\cite{wei2019semi} & NIQE / BRISQUE & \textbf{4.28/27.17} & 3.96 / 25.30 &  3.83 / 24.94  & 3.81/24.88 & 3.80/25.16 & 4.01 / 26.75  & \textbf{3.64 / 22.87}     \\ \hline
			\end{tabular}
			
		}
	\end{center}
	\vspace{-1em}
\end{table*}

The final loss function is defined as: 
\begin{equation}
\label{eq:final_loss}
\begin{aligned}
{L}_{f} &=  L^{cyc} + \lambda_p L^{p},\\
L^{cyc} &= |I_w - \hat{I}_w|_1 + |I_c - \hat{I}_c|_1 + L_{fwd}^{adv} + L_{rev}^{adv} + L_{identity},\\
L^p &= L_{fwd}^{p} + L_{rev}^{p},
\end{aligned}
\end{equation}
where $L^{cyc}$ is the CycleGAN loss,  $L_{fwd}^{adv}$ is the adversarial loss for the forward mapping, $L_{rev}^{adv}$ is the adversarial loss for the reverse mapping $L^p$ is the loss from the pseudo-labels, and $L_{identity}$ is identity loss, \ie~$L_{identity} = L_1(\mathcal{F}_{C\rightarrow W}(I_w),I_w)+L_1(\mathcal{F}_{W\rightarrow C}(I_c),I_c)$, $\lambda_p$ weights the contribution of loss from pseudo-label, and $ L_{fwd}^{p}$, $ L_{rev}^{p}$ are pseudo losses as described in Eq.~\ref{eq:lpfwd}, ~\ref{eq:lprev}.



\section{Experiments and results}

\subsection{Implementation details}


\noindent\textbf{Network architecture}: The base network is based on UNet \cite{ronneberger2015u} consisting of  Res2Net blocks \cite{gao2019res2net}.$^{\ref{ftn:dataset}}$

\noindent\textbf{Training}: The network  is trained using the Adam optimizer with a learning rate of 0.0002 and batch-size of 2 for a total of 60 epochs. We reduce the  learning rate by a factor of 0.5 at every 30 epochs. Note that the network is trained separately for every weather condition.

\noindent\textbf{Hyper-parameters}: We use $\lambda_p=0.03$. In Eq.~\ref{eq:final_mu}, using  all the vectors in $\mathbf{X}_c^z$ and $\mathbf{X}_c^s$ would lead to high computational and memory requirements. Instead, we use a subset of $N_n=32$ vectors which are nearest neighbors of $\tilde{\mathbf{z}}_c$. We use a 4-layer Deep GP. For kernel, we use a squared exponential kernel.  During training, the images are randomly cropped to  size  256$\times$256. In  all our experiments, we use $\frac{\beta}{\gamma}=1.0$. Ablation studies for kernels (heterogeneous/homogeneous), different values of $\lambda_p$, $N_n$, $L$ (no. of layers in Deep GP) can be found in the supplementary.

\vspace{-0.5em}
\subsection{Evaluation on real-world datasets}

As discussed in Section \ref{sec:intro}, the use of synthetic datasets for training the restoration networks does not necessarily result in optimal performance on real-world images. This can be attributed to the distribution gap between the synthetic   and   real-world images. Our approach is specifically designed to address this issue. To evaluate this, we evaluate and compare our approach with existing approaches for two tasks:  (i) restoration (de-raining/de-hazing) of  real-world images, (ii) evaluation of down-stream task performance on real-world images.  In these experiments, we train the existing fully-supervised approaches on a synthetic dataset, since they cannot exploit unpaired/unlabeled data. Similarly, in  the case of semi-supervised approaches, we use the synthetic data for fully supervised loss functions, and additionally unlabeled train split from a real-world dataset for the unlabeled loss functions. For the unpaired/unsupervised approaches, we use unpaired data from the train split of a real-world dataset.


\noindent\textbf{De-raining}: We conduct two experiments, where the networks are evaluated on two real-world datasets: SPANet  \cite{wang2019spatial}  and SIRR  \cite{wei2019semi}. In both cases, we use the DDN dataset-cite\cite{Authors17f}, which is a synthetic dataset, to train recent state-of-the-art fully-supervised approaches SPANet \cite{wang2019spatial},  PreNet \cite{ren2019progressive} and MSPFN-\cite{Zamir2021MPRNet}). For the semi-supervised approaches (SIRR \cite{wei2019semi}), we use  labeled data from  the DDN dataset  and unlabeled data from SPANet and SIRR datasets respectively for both the experiments. For the unpaired approaches, including ours, we use only unpaired data from SPANet and SIRR datasets respectively for both the experiments.  The results of the two experiments on the real-world datasets are shown in Table \ref{tab:real_spanet} and \ref{tab:real} respectively. For the evaluation on SPANet (Table \ref{tab:real_spanet}), we use PSNR/SSIM metrics since we have access to ground-truth for the test dataset. However, in the case of SIRR dataset (Table \ref{tab:real}), due to the unavailability of ground-truth on the test set, we use NIQUE/BRISUE scores, which are no-reference quality metrics. 

We make the following observations from Table \ref{tab:real_spanet}: (i) The results from the fully-supervised methods which are trained on synthetic dataset are sub-optimal as compared to the oracle\footnote{Oracle indicates the performance when the network is trained with full supervision using labels on the unlabeled data as well.} performance. This can be attributed to the domain shift problem as explained earlier. (ii) The semi-supervised approaches perform better as compared to the full-supervised methods, indicating that they are able to leverage unlabeled data. However, the gap with respect to oracle performance is still high which suggests that these methods are not able to exploit the real-world data completely, and are still biased towards the synthetic data. (iii) Our approach is able to not only outperform existing approaches by a significant margin but also minimizes the gap with respect to oracle performance. This demonstrates the effectiveness of the proposed Deep-GP based loss functions.  

\begin{figure*}[htp!]
	\begin{center}
		\includegraphics[width=0.19\linewidth,height=0.14\linewidth]{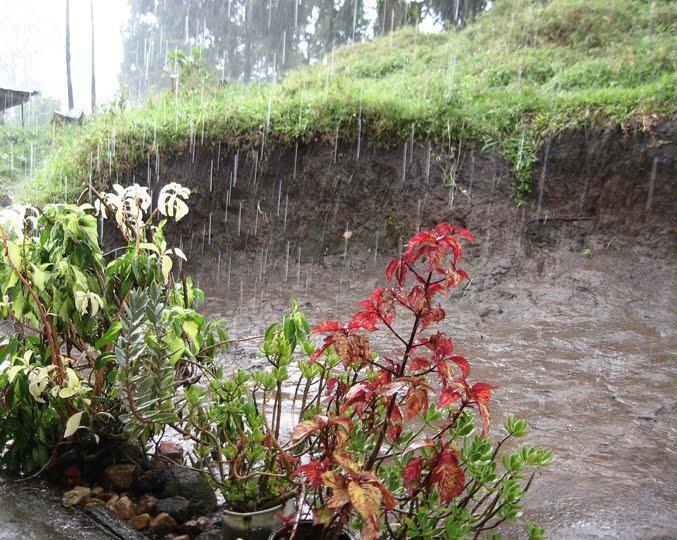}
		\includegraphics[width=0.19\linewidth,height=0.14\linewidth]{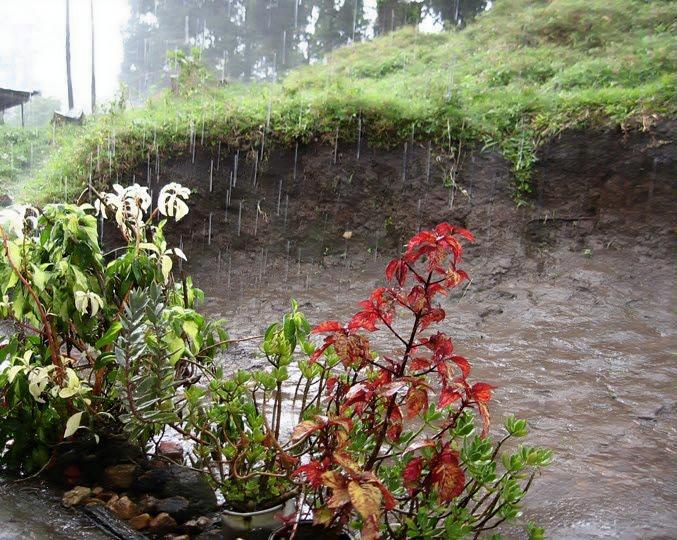}
		\includegraphics[width=0.19\linewidth,height=0.14\linewidth]{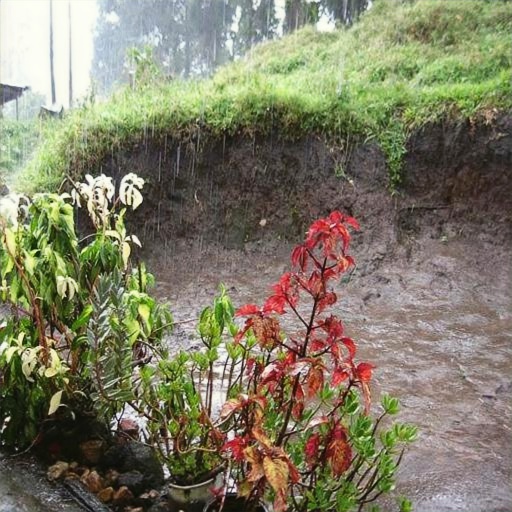}
		\includegraphics[width=0.19\linewidth,height=0.14\linewidth]{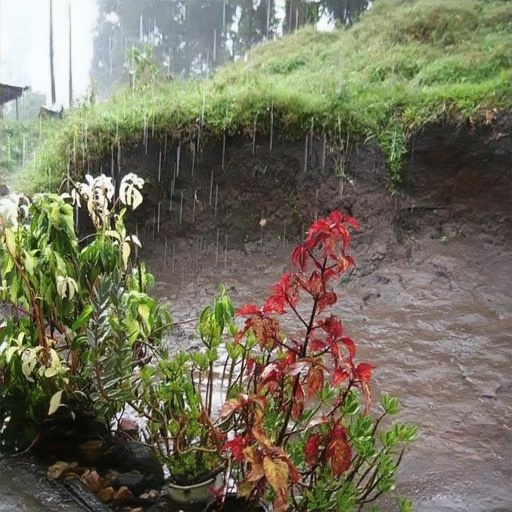}
		\includegraphics[width=0.19\linewidth,height=0.14\linewidth]{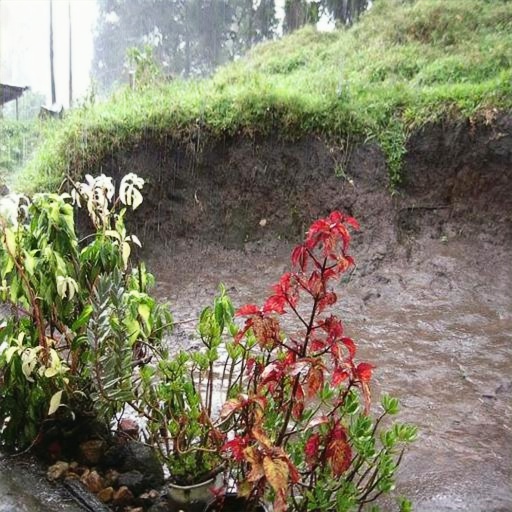}\\ \vskip2pt
		\includegraphics[width=0.19\linewidth,height=0.14\linewidth]{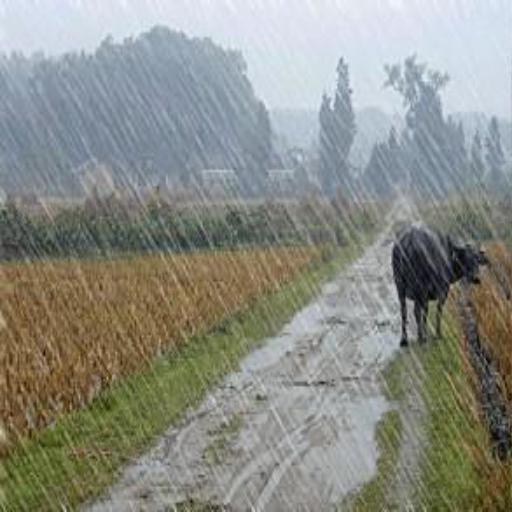}
		\includegraphics[width=0.19\linewidth,height=0.14\linewidth]{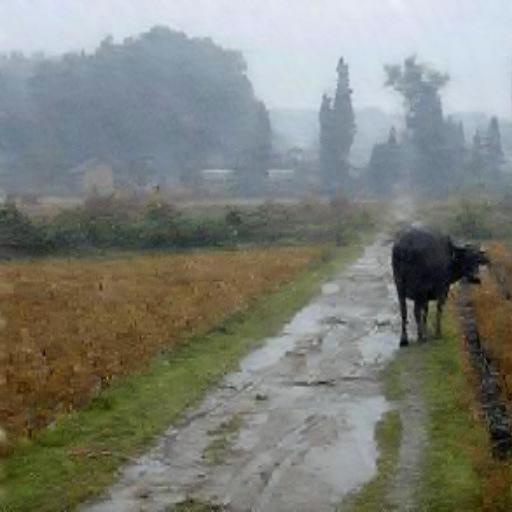}
		\includegraphics[width=0.19\linewidth,height=0.14\linewidth]{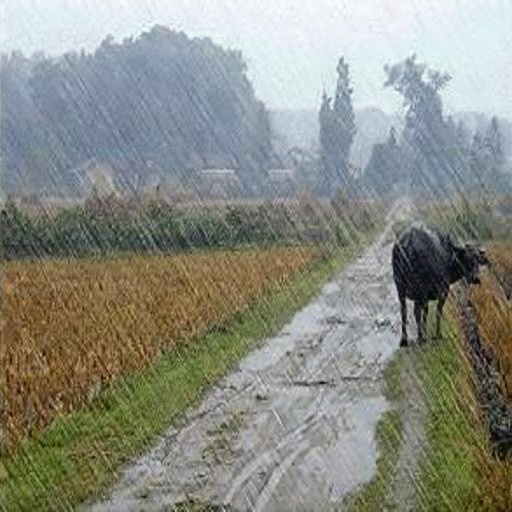}
		\includegraphics[width=0.19\linewidth,height=0.14\linewidth]{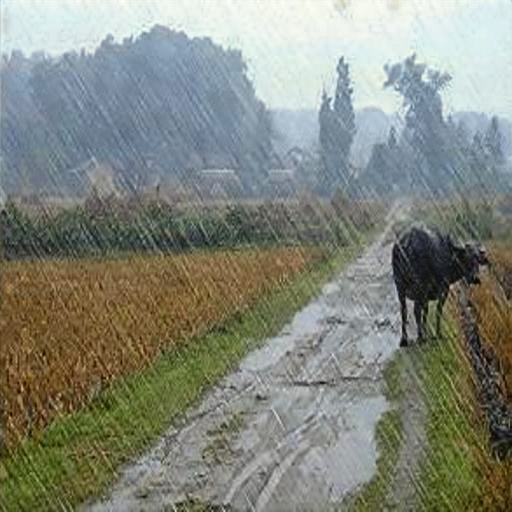}
		\includegraphics[width=0.19\linewidth,height=0.14\linewidth]{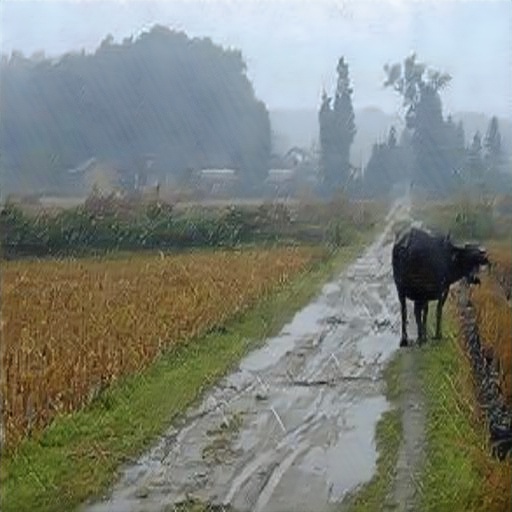}\\ \vskip2pt
		
	\end{center}
	\begin{flushleft}
		\vskip-10pt
		\hskip45pt Input\hskip65pt SPANet\cite{wang2019spatial}\hskip60pt PreNet\cite{ren2019progressive}\hskip45pt CycGAN\cite{zhu2017unpaired}   \hskip40pt DGP-CycGAN 
	\end{flushleft}
	\vskip -15pt \caption{Sample qualitative visualizations on real rainy images provided by authors of \cite{Authors18,wei2019semi}. Our results appear visually superior to fully-supervised results and the ground-truth.}
	\label{fig:results_real_rain}
\end{figure*}

Similar observations can be made for the evaluations on the SIRR dataset (see Table \ref{tab:real}). Further, these observations can also be made visually using  the qualitative results  shown in Fig.~\ref{fig:results_real_rain}


\noindent\textbf{De-hazing}:  We compare  the proposed method with recent fully-supervised SOTA methods (Grid-DeHaze \cite{liu2019griddehazenet} and EPDN \cite{qu2019enhanced}) on the RTTS \cite{li2019benchmarking} real-world dataset. To
train Grid-DeHaze and EPDN, we use paired  samples from the  SOTS-O~\cite{li2019benchmarking}  dataset.  To train   our method,   we use real-world unpaired samples from the RTTS dataset (train-set).

The results for these  experiments are shown in Table \ref{tab:real_haze}. Since we do not have access to   ground-truth data for real-world datasets, we use no-reference quality metrics (NIQE \cite{mittal2012making} and BRISQUE \cite{mittal2012no})  for comparison.  It can be observed that  the proposed method is able to achieve considerably better scores as compared to the fully-supervised networks, thus indicating that effectiveness of our approach.  Similar observations can be made for qualitative results as shown in Fig.~\ref{fig:results_real_hazy}.  In the supplementary material, we provide the comparisons the proposed method for the purpose of object detection using Cityscapes \cite{sakaridis2018semantic} (clean images) and RTTS \cite{li2019benchmarking} (hazy images).

\begin{table}[t!]	
	\caption{Results for de-hazing on real-world dataset (RTTS\cite{li2019benchmarking}). Lower numbers indicate better performance.}
	\label{tab:real_haze}
	\vskip -10pt
	\huge
	\renewcommand{\arraystretch}{1.1} 
	\begin{center}
		\resizebox{1\textwidth}{!}{
			\begin{tabular}{lcccc}
				\hline
				\textbf{Metric} & \textbf{Grid-DeHaze}\cite{liu2019griddehazenet} & \textbf{EPDN}\cite{qu2019enhanced} & \textbf{Cyc-GAN}\cite{zhu2017unpaired} & \textbf{Ours}\\
				\hline
				NIQE / BRISQUE &   29.27/46.80 & 29.75/47.09  & 30.16/47.38  & \textbf{28.19/44.48 }    \\ \hline
			\end{tabular}
			
		}
	\end{center}
	\vspace{-1em}
\end{table}

\begin{figure*}[htp!]
	\begin{center}
		\includegraphics[width=0.19\linewidth,height=0.14\linewidth]{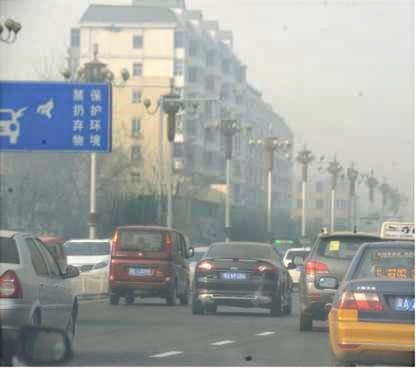}
		\includegraphics[width=0.19\linewidth,height=0.14\linewidth]{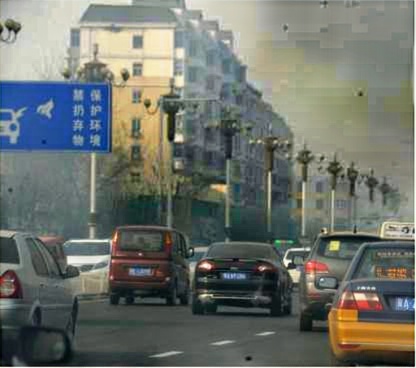}
		\includegraphics[width=0.19\linewidth,height=0.14\linewidth]{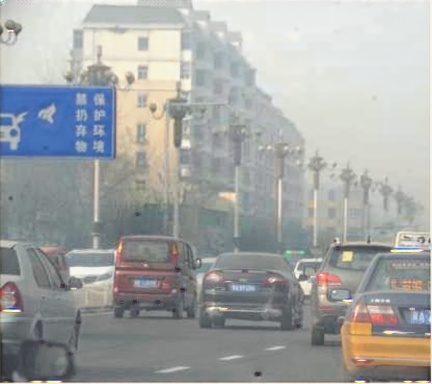}
		\includegraphics[width=0.19\linewidth,height=0.14\linewidth]{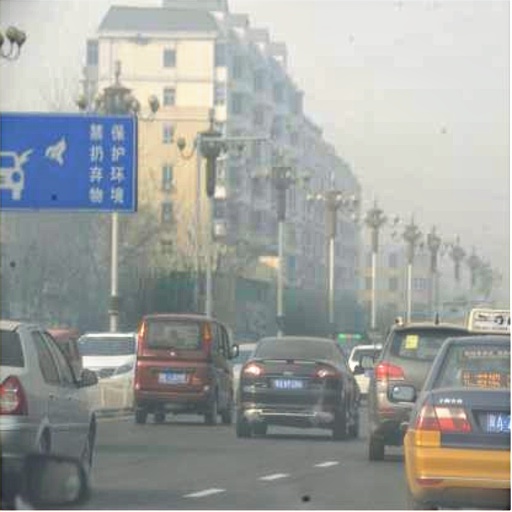}
		\includegraphics[width=0.19\linewidth,height=0.14\linewidth]{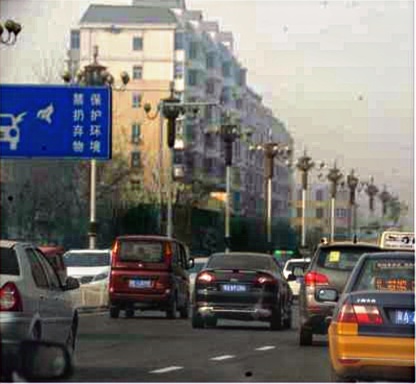}
		\\ \vskip2pt
		\includegraphics[width=0.19\linewidth,height=0.14\linewidth]{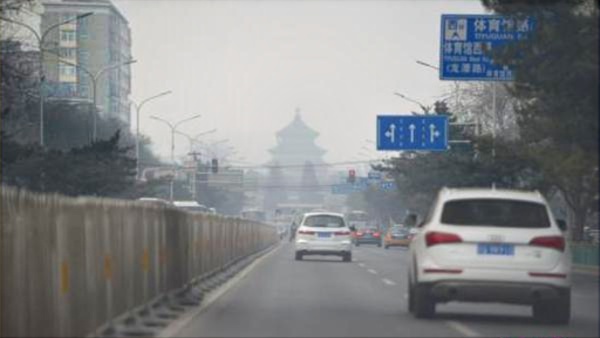}
		\includegraphics[width=0.19\linewidth,height=0.14\linewidth]{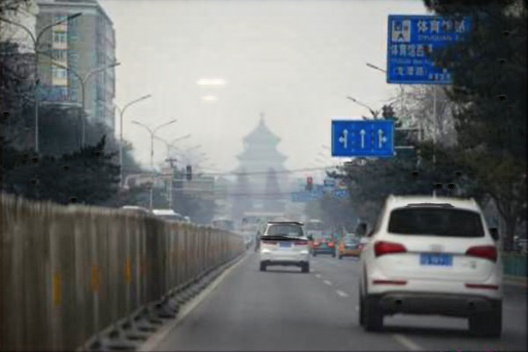}
		\includegraphics[width=0.19\linewidth,height=0.14\linewidth]{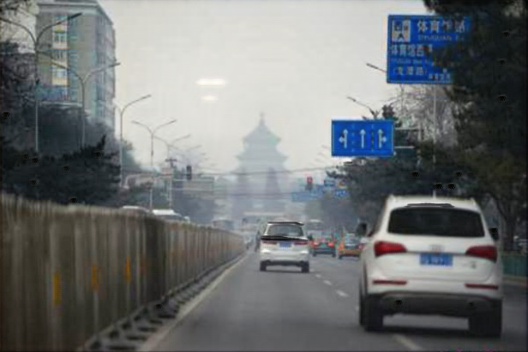}
		\includegraphics[width=0.19\linewidth,height=0.14\linewidth]{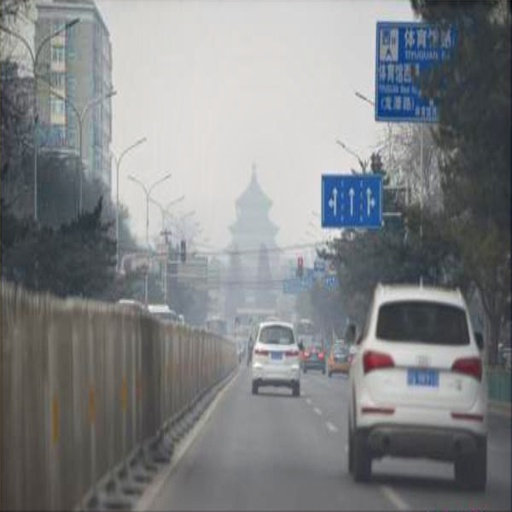}
		\includegraphics[width=0.19\linewidth,height=0.14\linewidth]{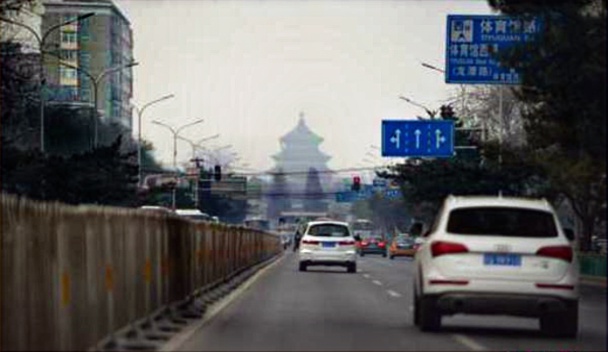}
		\\ \vskip2pt
	\end{center}
	\begin{flushleft}
		\vskip-10pt
		\hskip45pt Input\hskip70pt EPDN\cite{qu2019enhanced}\hskip45pt Grid-DeHaze\cite{liu2019griddehazenet}\hskip40pt CycGAN\cite{zhu2017unpaired}   \hskip40pt DGP-CycGAN 
	\end{flushleft}
	\vskip -15pt \caption{Sample qualitative visualizations on real hazy images of RTTS \cite{li2019benchmarking} dataset. Our results appear visually superior to fully-supervised results and the ground-truth.}
	\label{fig:results_real_hazy}
\end{figure*}
\vspace{-1em}

\subsection{Evaluation on synthetic datasets}

In order to verify the effectiveness of the proposed method, we conducted experiments using synthetic datasets, due to space constrain we provide the comparisons for  De-raining, and De-hazing tasks in supplementary material.  Here we provide the comparisons for De-snowing task.

\begin{table}[h!]
	\caption{Results for de-snowing. C: Classical, S: Supervised, U: Unsupervised. Metrics: PSNR (dB)$|$ SSIM.}
	\label{tab:desnow}
	\vskip-10pt
	\renewcommand{\arraystretch}{1.1} 
	\resizebox{0.8\textwidth}{!}{
		\begin{tabular}{llc}
			\Xhline{2\arrayrulewidth} 
			&                         & \textbf{Dataset}   \\ \hline
			\multirow{2}{*}{\textbf{Type}} & \multirow{2}{*}{\textbf{Method}} & \multirow{2}{*}{\textbf{Snow100k}\cite{liu2018desnownet}} \\ 
			&                                      &            \\ \hline
			\multirow{3}{*}{S}                                              & DerainNet\cite{Authors17d} (TIP'17)              & 22.8           $|$ 0.82          \\ 
			& DehazeNet\cite{cai2016dehazenet} (TIP'16)              & 23.9           $|$ 0.85          \\ 
			& DeSnowNet\cite{liu2018desnownet} (TIP'18)              & 30.1           $|$ 0.93          \\ \hline
			\multirow{2}{*}{U}                                                  & CycleGAN \cite{zhu2017unpaired} (ICCV'17)               & 23.5           $|$ 0.82          \\ 
			& DGP-CycleGAN(ours)  & \textbf{28.4  }    $|$  \textbf{0.88    }      \\ \cline{1-3} 
			& Oracle                  & 29.6           $|$ 0.91          \\
			\Xhline{2\arrayrulewidth} 
		\end{tabular}
	}
	\vspace*{-4mm}
\end{table}
The quantitative results along with comparison to other methods  for the desnowing task is shown in \ref{tab:desnow}. For a better understanding, we present results across categories of approaches : classical (C), fully-supervised (S) state-of-the-art (SOTA) methods and unsupervised (U). Note that our approach falls in the unsupervised category. For a fair comparison, we also present the results of fully-supervised training of our base network (``oracle'') which indicates the empirical upper-bound on the performance. We use two standard metrics for evaluation: peak signal-to-noise ratio (PSNR) and structural similarity index (SSIM). 

\noindent\textbf{De-snowing}: For the task of de-snowing, we perform the experiments on Snow100k \cite{liu2018desnownet} which consists of $10^5$ synthesized and 1329 real-world snowy images. The results are shown in Table \ref{tab:desnow}. Similar to the other two tasks, our method performs significantly better than CycleGAN while being comparable to the oracle and SOTA supervised techniques (DerainNet \cite{Authors17d}, DehazeNet \cite{cai2016dehazenet} and DeSnowNet \cite{liu2018desnownet}).

\section{Conclusion}

In this work, we presented a new approach for learning to restore images with weather-degradations using unpaired data. We build on a recent unpaired translation method (CycleGAN). Specifically, we derive new losses by modeling joint distribution of latent space vectors using Deep Gaussian Processes. The new losses enable learning of more accurate restoration functions as compared to the original CycleGAN. The proposed method (DGP-CycleGAN) is not weather-specific and  achieves high-quality restoration on multiple tasks like de-raining, de-hazing and de-snowing. Furthermore,  it can be effectively applied to other unpaired translation approaches like UNIT GAN. We also show that our approach  enhances performance of down-stream tasks like object detection. 
\section{Acknowledgements}
This work was supported by NSF CAREER award 2045489.
%

	\bibliographystyle{IEEEtran}
	\bibliography{egbib}

\end{document}


%
\title{Supplementary for Unsupervised Restoration of Weather-affected Images using Deep Gaussian Process-based CycleGAN}

	\author{Rajeev Yasarla\thanks{equal contribution} \qquad Vishwanath A. Sindagi \qquad Vishal M. Patel\\
		Johns Hopkins University\\
		Department of Electrical and Computer Engineering, Baltimore, MD 21218, USA\\
		{\tt\small \{ryasarl1, vishwanathsindagi, vpatel36\}@jhu.edu}
	}


%


\maketitle

\section{Introduction}
This is the supplementary material for the paper titled Unsupervised Restoration of Weather-affected Images using Deep Gaussian Process-based CycleGAN. We provide addiitional 
ablation studies, network architecture details, algorithm for proposed DGP-CycleGAN, details of ${L}^p_{rev}$, dataset details, and visualizations for qualitative analysis.

\section{Experiments}
\subsection{Evaluation on synthetic datasets}

In order to verify the effectiveness of the proposed method, we conducted experiments on 3 different weather-restoration tasks: (i) De-raining, and (ii) De-hazing, We use two standard metrics for evaluation: peak signal-to-noise ratio (PSNR) and structural similarity index (SSIM). 

The quantitative results along with comparison to other methods  for the three tasks are shown in Table~\ref{tab:derain}, \ref{tab:dehaze} and \ref{tab:desnow}. For a better understanding, we present results across categories of approaches : classical (C), fully-supervised (S) state-of-the-art (SOTA) methods and unsupervised (U). Note that our approach falls in the unsupervised category. For a fair comparison, we also present the results of fully-supervised training of our base network (``oracle'') which indicates the empirical upper-bound on the performance.

\noindent\textbf{De-raining}: For quantitative comparisons (see Table~\ref{tab:derain}), we include two classical approaches (DSC \cite{luo2015removing}, GMM \cite{Authors16}) and two supervised SOTA methods (SPANet \cite{wang2019spatial}, PReNet\cite{ren2019progressive}) from CVPR' 19. In the unsupervised category, we compare with CycleGAN \cite{zhu2017unpaired} and Derain-CycleGAN \cite{wei2019deraincyclegan} (Arxiv '19). The evaluations are performed on four datasets: Rain100L \cite{yang2017deep}, Rain100H \cite{yang2017deep}, DIDMDN \cite{Authors18}, SPANet \cite{wang2019spatial}. We  make the following observations:

In the unsupervised category, the proposed approach (DGP-CycleGAN) consistently achieves better results as compared to CycleGAN across all the datasets and metrics. For example, it obtains a minimum  improvement of 1.43 dB in terms of PSNR (Rain100H), while achieving 2.6 dB gain in some cases (DIDMDN). This signifies that the new losses ($L^p_{fwd}$,$L^p_{rev}$)  enable us to learn more accurate restoration functions.     Further, DGP-CycleGAN outperforms a recent unsupervised technique ,Derain-CycleGAN \cite{wei2019deraincyclegan}, in spite of the fact that Derain-CycleGAN uses rain-specific characteristics. Note that  our method is weather-agnostic.

Finally, the results from DGP-CycleGAN are within a reasonable range (maximum of 1.6 dB) as compared to the oracle. Similarly, the performance of our method is comparable to recent SOTA approaches (SPANet and PReNet) in most of the datasets (Rain100H, DIDMDN and SPANet).

\noindent\textbf{De-hazing}: As shown in Table~\ref{tab:dehaze},  we  compare our method with  two classical approaches (DCP \cite{he2010single}, NLD \cite{berman2016non}), three supervised SOTA methods (AOD-Net \cite{li2017aod}, GFN \cite{ren2018gated}, Grid-Dehaze\cite{liu2019griddehazenet}), and three unsupervised techniques (Golts \etal \cite{Golts2018UnsupervisedSI}, CycleGAN \cite{zhu2017unpaired} and CDNet \cite{dudhane2019cdnet}). The evaluations are performed on Synthetic Objective Testing Set (SOTS) of RESIDE\cite{li2019benchmarking}:  SOTS-I (500 indoor images) and SOTS-O (500 outdoor images)$^{\ref{ftn:dataset}}$. Similar to de-raining, the proposed approach (DGP-CycleGAN) consistently achieves significant improvement in PSNR (3.8 dB) and SSIM (0.11) as compared to CycleGAN. In addition, it outperforms other unsupervised techniques by a large margin while being reasonably comparable to fully-supervised SOTA techniques like AOD-Net and GFN.

\vskip-10pt
\begin{table}[hp!]
	\caption{Results for de-raining. C: Classical, S: Supervised, U: Unsupervised. Metrics: PSNR (dB) $|$ SSIM.}
	\huge
	\vskip-10pt
	\label{tab:derain}
	\setlength{\tabcolsep}{8pt} 
	\renewcommand{\arraystretch}{1.1} 
	\resizebox{0.8\textwidth}{!}{
		\begin{tabular}{llccc}
			\Xhline{2\arrayrulewidth} 
			&          &                \multicolumn{3}{c}{\textbf{Datasets}}                                                                                                  \\ \hline
			\multirow{2}{*}{\textbf{Type}} & \multirow{2}{*}{\textbf{Method}} & \multirow{2}{*}{\textbf{Rain100L}\cite{yang2017deep}} & \multirow{2}{*}{\textbf{Rain100H}\cite{yang2017deep}} & \multirow{2}{*}{\textbf{DIDMDN}\cite{Authors18}} \\ 
			&                         &           &           &                        \\ \hline 
			\multirow{2}{*}{C}                                                    & DSC\cite{luo2015removing} (ICCV'15)            & 27.3          $|$ 0.85          & 20.5          $|$  0.76          & 21.4        $|$  0.79         \\ 
			& GMM\cite{Authors16} (CVPR'16)            & 29.1          $|$ 0.87          & 21.7          $|$ 0.79          & 22.8         $|$ 0.81            \\ \hline 
			\multirow{3}{*}{S}                                               & SPANet\cite{wang2019spatial} (CVPR'19)          & 34.9          $|$ 0.97      &     26.3          $|$ 0.87        & 30.1        $|$  0.91              \\ 
			& PReNet\cite{ren2019progressive} (CVPR'19)         & 37.4          $|$ 0.97          & 26.8          $|$ 0.86          & 31.4         $|$ 0.91            \\  
			& MSPFN\cite{Kui_2020_CVPR} (CVPR'20)         & 37.2          $|$ 0.97          & 28.6          $|$ 0.87          & 32.8        $|$ 0.91            \\ \hline 
			\multirow{3}{*}{U}                                                   & Derain-CycleGAN\cite{wei2019deraincyclegan} (Arxiv '19) & 31.5          $|$  0.94          & --            $|$ --            & --           $|$ --         \\ 
			& CycleGAN\cite{zhu2017unpaired} (ICCV'17)       & 30.2          $|$  0.85          & 22.9          $|$ 0.74          & 25.8         $|$ 0.80            \\ 
			& DGP-CycleGAN(ours)  & 34.8          $|$  0.94         & 25.7         $|$  0.83          & 30.1         $|$ 0.88                \\ \cline{1-5}   
			& Oracle                  & 36.1          $|$  0.96          & 26.9          $|$  0.86          & 31.2         $|$ 0.90                \\ \Xhline{2\arrayrulewidth} 
		\end{tabular}
	}
\end{table}

\begin{table}[h!]
	\caption{Results for de-hazing. C: Classical, S: Supervised, U: Unsupervised. Metrics: PSNR (dB)$|$ SSIM.}
	\huge
	\label{tab:dehaze}
	\vskip-10pt
	\renewcommand{\arraystretch}{1.1} 
	\resizebox{0.8\textwidth}{!}{
		\begin{tabular}{llcc}
			\Xhline{2\arrayrulewidth} 
			&                         & \multicolumn{2}{c}{\textbf{Datasets}}                              \\ \hline
			\multirow{2}{*}{\textbf{Type}} & \multirow{2}{*}{\textbf{Method}} & \multirow{2}{*}{\textbf{SOTS-I}\cite{li2019benchmarking}} & \multirow{2}{*}{\textbf{SOTS-O}\cite{li2019benchmarking}} \\ 
			&                         &         &           \\ \hline
			\multirow{2}{*}{C}                                                    & DCP\cite{he2010single} (PAMI'10)                    & 16.6         $|$ 0.85         & 19.1          $|$ 0.86         \\ 
			& NLD\cite{berman2016non} (CVPR'16)                     & 17.3         $|$ 0.75         & 18.1          $|$ 0.80         \\ \hline
			\multirow{4}{*}{S}                                                     & AOD-Net\cite{li2017aod} (ICCV'17)                & 20.5        $|$ 0.82         & 24.1          $|$ 0.92         \\ 
			& GFN\cite{ren2018gated} (CVPR'19)                    & 24.9         $|$ 0.92         & 28.3          $|$ 0.93         \\ 
			& EPDN\cite{qu2019enhanced} (CVPR'19)            & 25.1        $|$ 0.92         & 22.6          $|$ 0.90    \\  
			& Grid-Dehaze\cite{liu2019griddehazenet} (ICCV'19)            & 32.2        $|$ 0.98         & 30.9          $|$ 0.96         \\ \hline
			\multirow{4}{*}{U}                                                   & Golts \etal\cite{Golts2018UnsupervisedSI} (TIP'19)                     & 19.3         $|$ 0.83         & 24.1          $|$ 0.93         \\ 
			& CDNet\cite{dudhane2019cdnet} (WACV'19)                     & 21.3        $|$ 0.89         & 22.9          $|$ 0.89         \\
			& CycleGAN\cite{zhu2017unpaired} (ICCV'17)               & 22.5         $|$ 0.79         & 21.6        $|$ 0.82         \\ 
			& DGP-CycleGAN(ours)  & 29.1        $|$ 0.93         & 27.9          $|$ 0.93         \\ \cline{1-4} 
			& Oracle                  & 31.3        $|$ 0.96         & 30.4          $|$ 0.95         \\ \Xhline{2\arrayrulewidth} 
		\end{tabular}
	}
\end{table}

\subsection{Experiments on downstream task}

In this experiment, we evaluate the proposed method for the purpose of object detection. We train the  restoration network using Cityscapes \cite{sakaridis2018semantic} (clean images) and RTTS \cite{li2019benchmarking} (hazy images). In order to asses the quality of restored (RTTS) images, we evaluate them on a down-stream task (object detection). We choose object detection since they are known to be  severely affected by domain gap arising from adverse weather conditions \cite{chen2018domain,saito2019strong}.  Specifically, we test the performance of Faster-RCNN \cite{ren2015faster} (which is  trained using Cityscapes \cite{sakaridis2018semantic}) on the restored images from (i) CycleGAN trained using Cityscapes and RTTS, and (ii) DGP-CycleGAN trained using Cityscapes and RTTS. The goal here is to improve the performance of a detection network (trained on a source dataset) on a target dataset without use of any target domain labels.  This problem is  known as unsupervised domain adaptive object detection \cite{chen2018domain}. 

Recently, works like DA-Faster \cite{chen2018domain}, SWDA \cite{saito2019strong}, and Sindagi \etal~\cite{sindagi2019prior} have attempted to address this issue by proposing different feature-level domain adaptation strategies. We follow a similar protocol as \cite{chen2018domain} for the evaluation purpose. Specifically, we use Cityscapes \cite{sakaridis2018semantic} as the source dataset and RTTS \cite{li2019benchmarking} as the target dataset. Note that RTTS dataset contains more than 4000 real world hazy images with object annotations. However, we do not use the annotations for training the network. We train the Faster-RCNN  network  on the source (Cityscapes) dataset by minimizing the bounding box and classification loss. As shown in Table \ref{Table:Object_detect}, the baseline Faster-RCNN achieves an overall mAP of 30.9. Further, we trained the recent feature-level adaptation techniques (DA-Faster, SWDA, and Sindagi \etal~\cite{sindagi2019prior}) by aligning the features of the target (RTTS) dataset and we observed improvements of  1.9, 2.6  and 3.2 mAP, respectively over the baseline. 

Another approach for addressing this problem is performing  pixel-level adaptation where we can apply I2I methods to translate from target to source before applying Faster-RCNN on the target images. An obvious technique is the CycleGAN method, and we obtained an overall mAP of 33.7 with this approach. Finally, we also evaluated our method (DGP-CycleGAN) for the translation, and we achieved an overall mAP of 34.3. These experiments show that the proposed method is able to outperform the feature-level adaptation methods while being superior to CycleGAN. 

\begin{table}[t!]
	\caption{Results of object detection  experiment using unsupervised domain adaptation protocol. B: Baseline F: Feature level DA, P: Pixel level DA}
	\vskip -10 pt
	\label{Table:Object_detect}
	\setlength{\tabcolsep}{8pt}
	\renewcommand{\arraystretch}{1.1} 
	\huge
	\resizebox{1.0\textwidth}{!}{
		\begin{tabular}{llcccccc}
			\Xhline{2\arrayrulewidth} 
			\multicolumn{2}{c}{\textbf{Method}} & \textbf{prsn} & \textbf{car} & \textbf{bus} & \textbf{bike} & \textbf{bcycle} & \textbf{mAP} \\ \hline
			B & Hazy & 46.6 & 39.8 & 11.7 & 19.0 & 37.0 & 30.9 \\ \hline
			\multirow{3}{*}{F} & DAFaster\cite{chen2018domain} (CVPR '18) & 37.7 & 48.0 & 14.0 & 27.9 & 36.0 & 32.8 \\ 
			& SWDA\cite{saito2019strong} (CVPR '19)& 42.0 & 46.9 & 15.8 & 25.3 & 37.8 & 33.5 \\
			& Sindagi \etal \cite{sindagi2019prior} (ECCV '20) & 37.4 & 54.7 & 17.2 & 22.5 & 38.5 & 34.1 \\ \hline
			\multirow{2}{*}{P} & CycleGAN\cite{zhu2017unpaired} (ICCV '17) & 56.3 & 47.8 & 19.3 & 27.3 & 18.0 & 33.7 \\ 
			& DGP-CycleGAN (ours) & 56.7 & 48.9 & 19.8 & 27.6 & 18.5 & 34.3 \\ \Xhline{2\arrayrulewidth} 
	\end{tabular}}
\end{table}

\subsection{Generalization to other I2I approaches}
\label{ssec:unitgan}
We verify the effectiveness of the proposed method of employing losses based on pseudo-labels for other image-to-image translation (I2I) approaches. Specifically, we conduct experiments with UNIT GAN \cite{liu2017unsupervised} for de-raining  on two datasets: Rain100H \cite{yang2017deep} and DIDMDN \cite{Authors18} and the results are shown in Table~\ref{tab:generalization}. Note that  employing Deep Gaussian Processes to introduce losses into the UNIT framework results in significant improvements in both the datasets.
\begin{table}[h!]
	\caption{Experiments on generalization (to other translation approach UNIT \cite{liu2017unsupervised}). Metrics: PSNR (dB)$|$ SSIM. }
	\label{tab:generalization}
	\vskip-8pt
	\resizebox{0.67\textwidth}{!}{
		\begin{tabular}{llcc}
			\Xhline{2\arrayrulewidth}
			\textbf{Method}                 &\textbf{Rain100H}\cite{yang2017deep} & \textbf{DIDMDN} \cite{Authors18}\\ 
			\hline
			UNIT  \cite{liu2017unsupervised}             & 20.8  $|$ 0.71          & 24.3     $|$        0.77          \\ 
			DGP-UNIT & 23.6            $|$ 0.80            & 26.5            $|$ 0.83          \\ \Xhline{2\arrayrulewidth}
		\end{tabular}
	}
\end{table}
\subsection{CycleGAN BaseNetwork experiments}
We use the same BaseNetwork for CycGAN~\cite{zhu2017unpaired} and DGP-CycGAN (ours) as shown in the Figure~\ref{fig:overview}. We construct BaseNet based on UNet~\cite{ronneberger2015u} and Res2Block~\cite{gao2019res2net} as shown in the Figure~\ref{fig:overview}.  Details of this BaseNet are given in the section~\ref{sec:net_arch}.  As shown in the Table~\ref{tab:cyc_basenet}, by using BaseNet in the CycGAN~\cite{zhu2017unpaired} we obtained higher performance when compare using UNet~\cite{ronneberger2015u} in the CycGAN~\cite{zhu2017unpaired}.

\begin{table}[h!]
	\caption{Experiments on base network for CycGAN~\cite{zhu2017unpaired} and DGP-CycGAN. Metrics: PSNR (dB)$|$ SSIM.}
	\label{tab:cyc_basenet}
	\vskip-8pt
	\resizebox{\textwidth}{!}{
		\begin{tabular}{llcc}
			\Xhline{2\arrayrulewidth}
			\textbf{Method}                 & CycGAN w/ UNet & CycGAN w/ BaseNet\\ 
			\hline
			SPANet  \cite{wang2019spatial}             & 30.6  $|$ 0.84          & 32.4     $|$        0.86          \\ 
			\Xhline{2\arrayrulewidth}
		\end{tabular}
	}
\end{table}

\subsection{ MSFPN experiments}
For the experiments in the section 4.2.1 (Restoration of real-world image) of the main paper we train fully-supervised approaches  SPANet~\cite{wang2019spatial}, PreNet~\cite{ren2019progressive}, MSFPN~\cite{Kui_2020_CVPR} using DDN~\cite{Authors17f} dataset. We choose DDN~\cite{Authors17f} dataset for the experiments in the section 4.2.1 of the main paper since MSFPN~\cite{Kui_2020_CVPR} performed better on SPANet test set using DDN~\cite{Authors17f} dataset as shown in the Table~\ref{tab:msfpn}.

\begin{table}[h!]
	\caption{MSFPN experiments. Metrics: PSNR (dB)$|$ SSIM. }
	\label{tab:msfpn}
	\vskip-8pt
	\resizebox{\textwidth}{!}{
		\begin{tabular}{llcccc}
			\Xhline{2\arrayrulewidth}
			\textbf{Method}                 & DIDMDN~\cite{Authors18} & RAIN100L-100H~\cite{yang2017deep}  & DDN~\cite{Authors17f}\\ 
			\hline
			SPANet  \cite{wang2019spatial}              & 31.7  $|$ 0.90         & 32.1    $|$        0.91   &  33.8    $|$        0.93        \\ 
			\Xhline{2\arrayrulewidth}
		\end{tabular}
	}
\end{table}
\section{Ablation analysis}

In this section, we provide ablation studies for kernels (heterogeneous/homogeneous), different values of $\lambda_p$, $N_n$, $L$ (no. of layers in Deep GP).

\subsection{Number of hidden layers}
We conducted experiments on Rain100H\cite{yang2017deep}, DIDMDN\cite{Authors18}, SOTS-Out\cite{li2019benchmarking} for different number of hidden layers L in Deep Gaussian Processes. The results of this experiment are tabulated in Table~\ref{Table:L}. It can be clearly observed that increasing the number of layers results in better representational power, and hence improved results. 
\begin{table}[ht!]
	\caption{Ablation study with different number of layers}
	\label{Table:L}
	\renewcommand{\arraystretch}{1.25} 
	\huge
	\resizebox{\textwidth}{!}{
		\begin{tabular}{llcccc}
			\Xhline{2\arrayrulewidth} 
			\multirow{2}{*}{\begin{tabular}[c]{@{}l@{}}\textbf{Weather}\\ \textbf{Condition}\end{tabular}} & \multirow{2}{*}{\textbf{Dataset}} & \multicolumn{4}{c}{\textbf{Number of hidden layers}}              \\ 
			&                          & \textbf{L=1}          & \textbf{L=2}          & \textbf{L=3}          & \textbf{L=4}          \\ \hline
			\multirow{2}{*}{Derain}                                                      & Rain100H\cite{yang2017deep}                 & 23.51 $|$ 0.79 & 23.78 $|$ 0.82 & 23.94 $|$ 0.82 & 24.20 $|$ 0.83 \\ 
			& DIDMDN\cite{Authors18}                   & 27.75 $|$ 0.84 & 27.96 $|$ 0.86 & 28.16 $|$ 0.86 & 28.38 $|$ 0.88 \\ \hline
			Dehaze                                                                       & SOTS-Out\cite{li2019benchmarking}                  & 23.47 $|$ 0.87 & 24.28  $|$ 0.90 & 24.95 $|$ 0.92 & 25.43 $|$ 0.93 \\ \Xhline{2\arrayrulewidth} 
		\end{tabular}
	}
\end{table}

\subsection{Different composition of kernels}
We conducted experiments involving different composition of heterogeneous and homogeneous kernels while formulating Deep Gaussian processes. Table~\ref{Table:kernels} shows performance of 2-layer Deep GP on Rain100H\cite{yang2017deep}, and DIDMDN\cite{Authors18} datasets using different composition of square exponential (SE[.]), linear (LIN[.]), and squared cosine (SC[.]) kernel functions.
\begin{table}[ht!]
	\caption{Ablation study with different composition of kernels}
	\label{Table:kernels}
	\renewcommand{\arraystretch}{1.25} 
	\resizebox{0.8\textwidth}{!}{
		\begin{tabular}{lccc}
			\Xhline{2\arrayrulewidth} 
			\textbf{Dataset} &\textbf{SE[SC]} & \textbf{SE[LIN]} & \textbf{SE[SE]} \\
			\hline
			Rain100H\cite{yang2017deep}                         &  23.3  $|$ 0.82                & 23.6              $|$ 0.82             & 23.8             $|$ 0.82             \\
			DIDMDN\cite{Authors18}                            &  27.5 $|$ 0.85                & 27.7              $|$ 0.86             & 28.0             $|$ 0.86             \\ \Xhline{2\arrayrulewidth} 
		\end{tabular}
	}
\end{table}
\subsection{Different values of $N_n$ and $\lambda_p$}
We conduct different experiments for various values of $N_n$ and $\lambda_p$ on Rain100H\cite{yang2017deep} and Rain100L\cite{yang2017deep} datasets. Table~\ref{Table:N_n} shows the performance of DGP-CycleGAN when we vary the number of nearest neighbors used in computation the pesudo label $\mathbf{\tilde{z}}_c^p$ or $\mathbf{\tilde{z}}_w^p$. Table~\ref{Table:lambda} shows the performance of DGP-CycleGAN for different vaules of $\lambda_p$ in the loss ${L}_{f}$ while training the DGP-CycleGAN.
\begin{table}[ht!]	
	\vskip-8pt
	\caption{Ablation study for number of nearest neighbors, $N_n$ }
	\huge
	\setlength{\tabcolsep}{8pt} 
	\renewcommand{\arraystretch}{1.25} 
	\resizebox{.71\textwidth}{!}{
		\label{Table:N_n}
		\begin{tabular}{lccc}
			\Xhline{2\arrayrulewidth} 
			\textbf{Dataset} &\textbf{$N_n=16$} & \textbf{$N_n=32$} & \textbf{$N_n=64$} \\
			\hline
			Rain100L\cite{yang2017deep}                          & 31.28                & 31.91            & 31.94            \\
			Rain100H\cite{yang2017deep}                             & 23.75               &    24.18         & 24.12           \\ \Xhline{2\arrayrulewidth} 
		\end{tabular}		
	}
\end{table}

\begin{table}[ht!]	
	\vskip-8pt
	\caption{Ablation study for different values of $\lambda_p$ }
	\huge
	\setlength{\tabcolsep}{8pt} 
	\renewcommand{\arraystretch}{1.25} 
	\resizebox{0.8\textwidth}{!}{
		\label{Table:lambda}
		\begin{tabular}{lccc}
			\Xhline{2\arrayrulewidth} 
			\textbf{Dataset} &\textbf{$\lambda_p=0.3$} & \textbf{$\lambda_p=0.03$} & \textbf{$\lambda_p=0.003$} \\
			\hline
			Rain100L\cite{yang2017deep}                          & 31.88                & 31.91            & 31.83           \\
			Rain100H\cite{yang2017deep}                            & 24.17               &    24.18         & 24.06           \\ \Xhline{2\arrayrulewidth} 
		\end{tabular}
		
	}
\end{table}
\begin{algorithm}
	\label{alg_train}
	\SetAlgoLined
	\KwIn{unpaired data, $\m{D}=\m{D}_w \cup \m{D}_c$, where $\m{D}_w=\{I_w^i\}_{i=1}^N$  and $\m{D}_c=\{I_c^i\}_{i=1}^N$, intial weights of forward mapping $\Theta_{W\rightarrow C}$, and reverse mapping $\Theta_{C\rightarrow W}$.}
	\KwResult{optimized network parameters of DGP-CycleGAN ($\Theta_{W\rightarrow C}$ and $\Theta_{C\rightarrow W}$)}
	\For{every epoch }{
		\For{$I_w \in\m{D}_w$}{
			$\tilde{I}_c,\: s_w,\: z_w$ = $\m{F}_{W\rightarrow C}(I_w,\Theta_{W\rightarrow C})$\\
			store $s_w,\: z_w$ $\rightarrow$ in $\mathbf{F}^{s}_{w},\mathbf{F}^{z}_{w}$ respectively
		}
		\For{$I_c \in\m{D}_c$}{
			$\tilde{I}_w,\: s_c,\: z_c$ = $\m{F}_{C\rightarrow W}(I_c,\Theta_{C\rightarrow W})$\\
			store $s_c,\: z_c$ $\rightarrow$ in $\mathbf{F}^{s}_{c},\mathbf{F}^{z}_{c}$ respectively
		}
		\For{$I_w,I_c \in\m{D}$}{
			$\tilde{I}_c,\: s_w,\: z_w$ = $\m{F}_{W\rightarrow C}(I_w,\Theta_{W\rightarrow C})$\\
			$\hat{I}_w,\: \tilde{s}_c,\: \tilde{z}_c$ = $\m{F}_{C\rightarrow W}(\tilde{I}_c,\Theta_{C\rightarrow W})$\\
			compute nearest neighbors of $\tilde{z}_c$ in $\mathbf{F}^{z}_{c}$ using distance metric\\
			compute pseudo-label $\mathbf{\tilde{z}}_c^p$, and $\mathbf{\tilde{\Sigma}}^{z}_{{c}}$ using Deep GP\\
			compute ${L}^{p}_{fwd}$\\
			$\tilde{I}_w,\: s_c,\: z_c$ = $\m{F}_{C\rightarrow W}(I_c,\Theta_{C\rightarrow W})$\\
			$\hat{I}_c,\: \tilde{s}_w,\: \tilde{z}_w$ = $\m{F}_{W\rightarrow C}(\tilde{I}_w,\Theta_{W\rightarrow C})$\\
			compute nearest neighbors of $\tilde{z}_w$ in $\mathbf{F}^{z}_{w}$ using distance metric\\
			ompute pseudo-label $\mathbf{\tilde{z}}_w^p$, and $\mathbf{\tilde{\Sigma}}^{z}_{{w}}$ using Deep GP\\
			compute ${L}^p_{rev}$\\
			compute cycle-GAN loss $L^{cyc}$\\
			update weights $\Theta_{W\rightarrow C}$ and $\Theta_{C\rightarrow W}$ using loss ${L}_{f}$
		}
	}
	
	\caption{Pseudo code for training DGP-CycleGAN.}
\end{algorithm}

\section{Network Architecture details }\label{sec:net_arch}
\begin{figure*}[ht!]
	\begin{center}
		\includegraphics[width=1\linewidth]{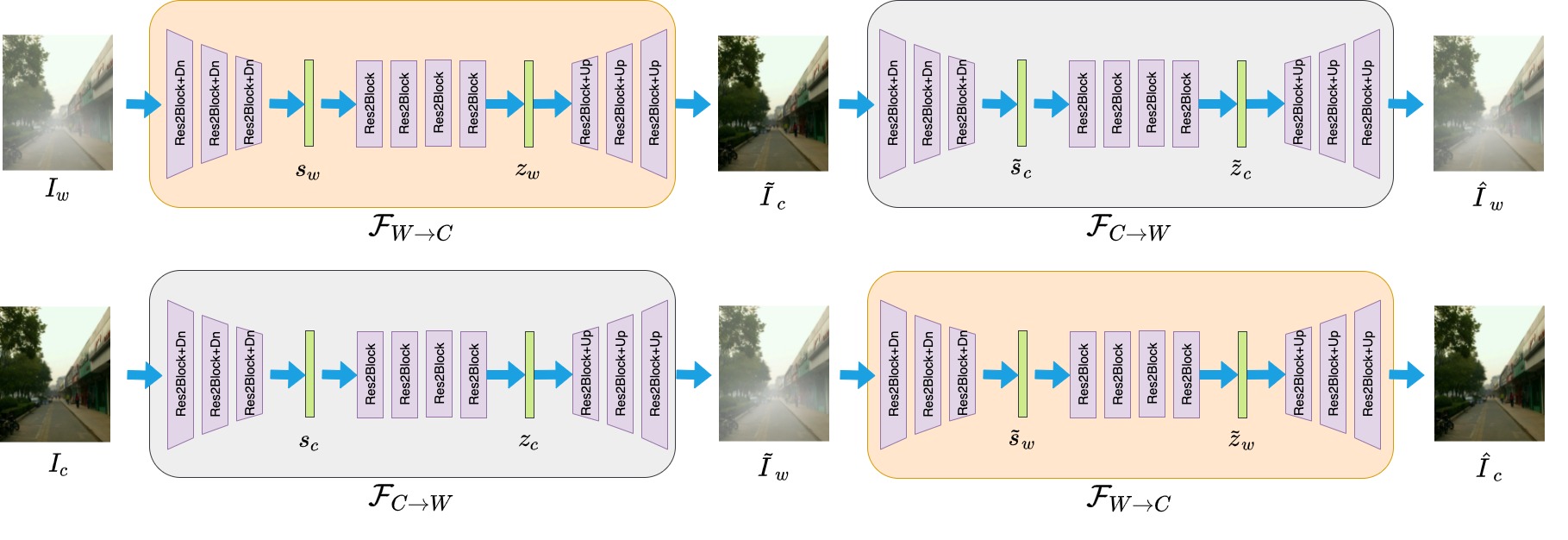}
	\end{center}
	\vskip -10pt 
	\caption{Overview of Network architecture of the proposed Deep Gaussian Processes-based CycleGAN. Dn means down-sample, and Up means Up-sample.}
	\label{fig:overview}
	
\end{figure*}

The backbone network of two generators of forward mapping $\m{F}_{W\rightarrow C}$ , and reverse mapping $\m{F}_{C\rightarrow W}$ are based on UNet \cite{ronneberger2015u} consisting of  Res2Net blocks \cite{gao2019res2net}. As shown in the Figure~\ref{fig:overview}, the base network consists of 10 Res2Net blocks \cite{gao2019res2net}.  Figure~\ref{fig:overview} shows in the forward mapping network ($\m{F}_{W\rightarrow C}$), a weather-degraded image $I_w$ is mapped to a latent embedding $\mathbf{s_w}$, which is then mapped  to another embedding $\mathbf{z_w}$, before being mapped  to restored (clean) image $\tilde{I}_c$. The restored (cleaned) image $\tilde{I}_c$ is then forwarded through the reverse mapping network ($\m{F}_{C\rightarrow W}$) to produce latent vectors $\mathbf{\tilde{s}_c}$ and $\mathbf{\tilde{z}_c}$, before being mapped to a reconstructed weather-degraded image $\hat{I}_w$. 

\section{Algorithm for proposed DGP-CycleGAN}
Algorithm 1 shows the detailed steps involved in training of DGP-CycleGAN.

\section{Derivation of ${L}^p_{rev}$}
A clean image $I_c$ in the reverse mapping network ($\m{F}_{C\rightarrow W}$) as shown in the Figure~\ref{fig:overview}, is mapped to a latent embedding $\mathbf{s_c}$, which is then mapped  to another embedding $\mathbf{z_c}$, before being mapped  to  weather-degraded image $\tilde{I}_w$. The generated weather-degraded image $\tilde{I}_w$ is then forwarded through the forward mapping network ($\m{F}_{W\rightarrow C}$) to produce latent vectors $\mathbf{\tilde{s}_w}$ and $\mathbf{\tilde{z}_w}$, before being mapped to a reconstructed clean image $\hat{I}_c$. 

As explained in the section 3 of our main paper, similarly we formulate a jonit distribution using Deep Gaussian processes with effective kernel function as follows, 
\begin{equation}
\label{eq:joint_dis}
\resizebox{1.0\hsize}{!}{
	$
	\begin{array}{c}
	\left[\begin{array}{c}
	\mathbf{z}_{w} \\
	\tilde{\mathbf{z}}_{w}
	\end{array}\right] \sim G P\left(\left[\begin{array}{l}
	\boldsymbol{\mu}^L_{w} \\
	\boldsymbol{\tilde{\mu}}^L_{{w}}
	\end{array}\right], \left[\begin{array}{cc}
	K_{eff}\left(\mathbf{s}_{w}, \mathbf{s}_{w}\right) & K_{eff}\left(\mathbf{s}_{w}, \mathbf{\tilde{s}}_{w}\right) \\
	K_{eff}\left(\mathbf{\tilde{s}}_{w},\mathbf{s}_{w}\right) & K_{eff}\left(\mathbf{\tilde{s}}_{w},\mathbf{\tilde{s}}_{w}\right)
	\end{array}\right]+\sigma_{\epsilon}^{2}\mathbb{I}\right),\\
	\end{array}
	$}
\end{equation}
where $\mathbb{I}$ is identity matrix, $\sigma_{\epsilon}^{2}$ is  the additive noise variance that is set to 0.01. By conditioning the above distribution, we obtain the following distribution for $\mathbf{z}^{p}_w$:
\begin{equation}
P\left(\mathbf{\tilde{z}}^{p}_{w} | \mathbf{X}^{z}_{w},\mathbf{X}^{s}_{w}, \mathbf{\tilde{s}}_{w}\right)=\mathcal{N}\left(\boldsymbol{\tilde{\mu}}^L_{w},\mathbf{\tilde{\Sigma}}_{{w}}\right),
\end{equation}
where, $\mathbf{X}^{s}_{w}$ and $\mathbf{X}^{z}_{w}$ are matrices of latent projections $\mathbf{s}_w$'s and $\mathbf{z}_w$'s, respectively of all weather-degraded images in the dataset, and $\boldsymbol{\tilde{\mu}}_{w},\mathbf{\tilde{\Sigma}}_{{w}}$ are defined as follows:
\begin{equation}
\label{eq:final_mu}
\resizebox{1.0\hsize}{!}{
	$
	\begin{array}{c}
	\boldsymbol{\tilde{\mu}}_{{{w}}}=K_{eff}\left(\mathbf{\tilde{s}}_{w}, \mathbf{X}^{s}_{w}\right)\left[K_{eff}\left(\mathbf{X}^{s}_{w}, \mathbf{X}^{s}_{w}\right)+\sigma_{\epsilon}^{2} I\right]^{-1} \mathbf{X}^{z}_{w}, \\
	\mathbf{\tilde{\Sigma}}^{z}_{{w}}=K_{eff}\left(\mathbf{\tilde{s}}_{w}, \mathbf{\tilde{s}}_{w}\right)-K_{eff}\left(\mathbf{\tilde{s}}_{w}, \mathbf{X}^{s}_{w}\right)\left[K_{eff}\left(\mathbf{X}^{s}_{w}, \mathbf{X}^{s}_{w}\right)+\sigma_{\epsilon}^{2} I\right] 
	K_{eff}\left(\mathbf{X}^{s}_{w}, \mathbf{\tilde{s}}_{w}\right)+\sigma_{\epsilon}^{2}\mathbb{I}.
	\end{array}
	$}
\end{equation}
Thus, we can use the expression for $\boldsymbol{\tilde{\mu}}^L_w$ from Eq.~\ref{eq:final_mu} as  $\mathbf{\tilde{z}}^{p}_w$. We then use $\mathbf{\tilde{z}}^{p}_w$ to supervise  $\mathbf{\tilde{z}_w}$ and define the loss function as follows:
\begin{equation}
\label{eq:lprev}
{L}^p_{rev} = \left(\mathbf{\tilde{z}}_{w}-\mathbf{\tilde{z}}^{p}_{w}\right)^{\textrm{T}}\left(\mathbf{\tilde{\Sigma}}^{z}_{{w}}\right)^{-1}\left(\mathbf{\tilde{z}}_{w}-\mathbf{\tilde{z}}^{p}_{w}\right) + \log\left(\det{|\mathbf{\tilde{\Sigma}}^{z}_{{w}}|}\right).
\end{equation}

To summarize, given the latent embedding vectors corresponding to  ``weather-degraded images'' from the first  and second intermediate layers ($\mathbf{s}_w$ and $\textbf{z}_w$, respectively), and latent embedding vectors corresponding to  ``generated weather-degraded images'' from the first intermediate layer $\mathbf{\tilde{s}}_w$, we obtain the pseudo-labels $\mathbf{\tilde{z}}_w^p$  using  Eq.~\ref{eq:final_mu}. These pseudo-labels are then used to supervise at the second intermediate layer in the network $\mathbf{z}_{w}$ as shown in the Algorithm~1.

\section{Dataset details}
\subsection{De-raining}
De-raining experiments are performed on four datasets: Rain100L \cite{yang2017deep}, Rain100H \cite{yang2017deep}, DIDMDN \cite{Authors18}, SPANet \cite{wang2019spatial}. Authors of \cite{yang2017deep} published two training and test sets, i.e Rain100L and Rain100H. Rain100L contains 1800 low rain training images and 100 low rain testing images.  Rain100H contains 1800 heavy rain training images and 100 heavy rain testing images. DIDMDN rain dataset \cite{Authors18}, contains 4000 low density rain, 4000  medium density rain, and 4000 high density rain images for training, and 1200 testing images ( 400 low density, 400  medium density, and 400 high density rain images). SPANet \cite{wang2019spatial} is biggest real rain dataset and, it contains 342 real rain image frames for training, and  1000 real rain images for testing.
\subsection{De-hazing}
We use the following datasets for the task of de-hazing: (i)Indoor training set of RESIDE~\cite{li2019benchmarking} that contains 13990 indoor hazy images, and 1399 clean images, and (ii) Outdoor training set of RESIDE~\cite{li2019benchmarking} that contains 313,950 outdoor hazy images, and 8477 clean images.  RESIDE~\cite{li2019benchmarking} contains Synthetic Objective Testing Set (SOTS) and Real-world Task driven Testing Set (RTTS). SOTS testing contains 500 indoor hazy testing images SOTS-I, and 500 outdoor hazy testing images SOTS-O. RTTS testing contains 4332 real world hazy images. For object detection experiment in section 4.3 of our main paper we use Cityscapes~\cite{sakaridis2018semantic} (clean images) and RTTS \cite{li2019benchmarking} (hazy images). In order to asses the quality of restored (RTTS) images, we evaluate them on a down-stream task (object detection). We use 2975 clean images from the dataset Cityscapes~\cite{sakaridis2018semantic}, and RTTS hazy images for object detection experiment in our main paper.
\subsection{De-snowing}
For the task of de-snowing, we perform the experiments on Snow100k \cite{liu2018desnownet} which consists of $10^5$ synthesized and 1329 real-world snowy images. Snow100k contains 50,000 training images, 50,000 testing images.
\section{Qualitative Analysis}
\begin{figure*}[htp!]
	\begin{center}
		\includegraphics[width=0.19\linewidth,height=0.14\linewidth]{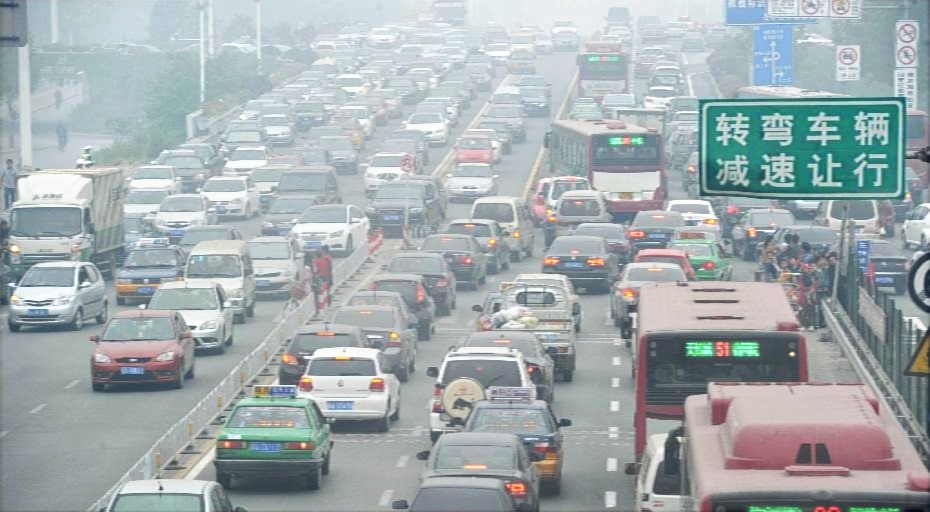}
		\includegraphics[width=0.19\linewidth,height=0.14\linewidth]{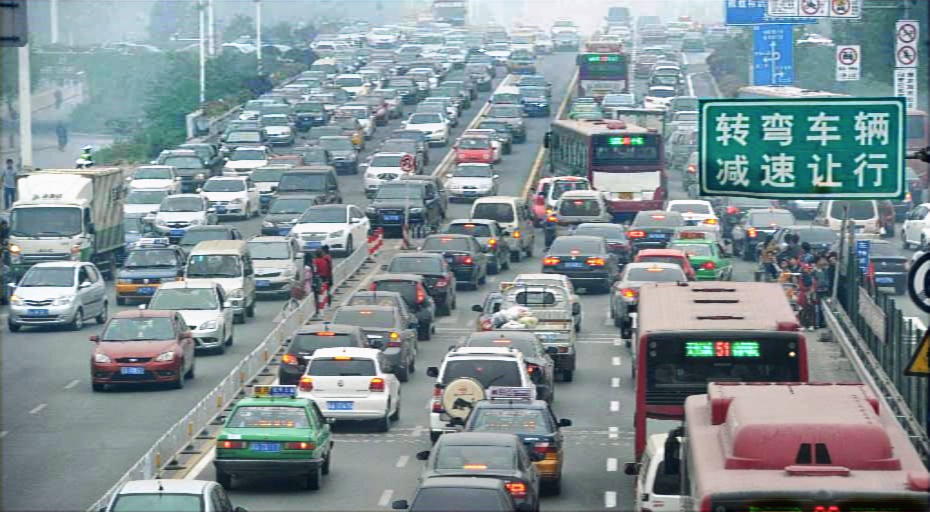}
		\includegraphics[width=0.19\linewidth,height=0.14\linewidth]{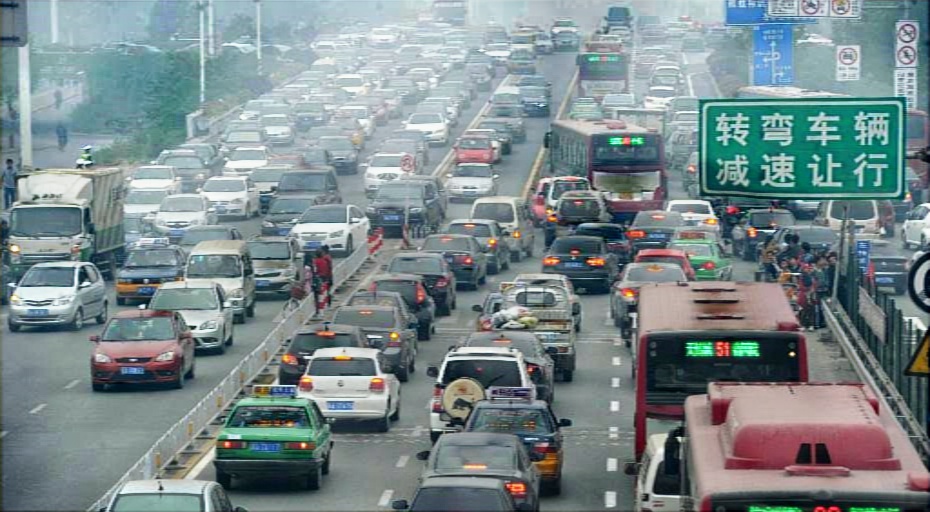}
		\includegraphics[width=0.19\linewidth,height=0.14\linewidth]{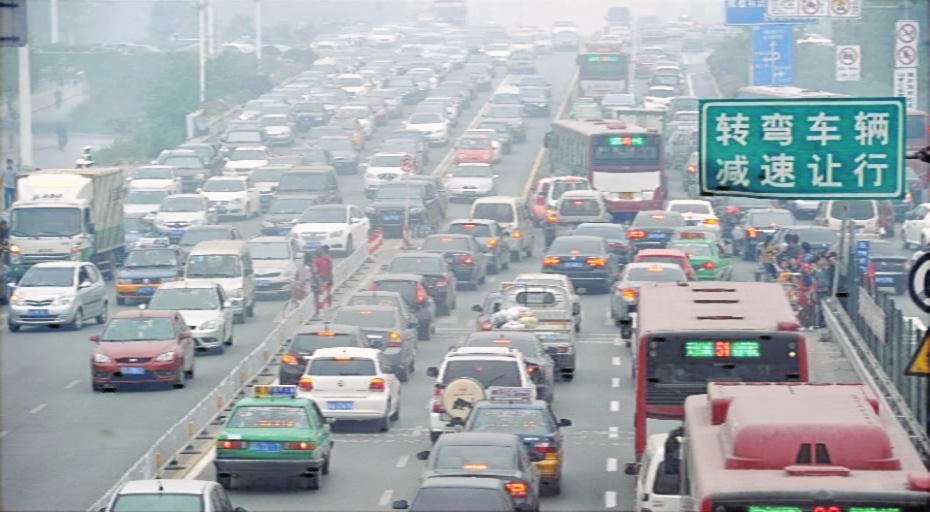}
		\includegraphics[width=0.19\linewidth,height=0.14\linewidth]{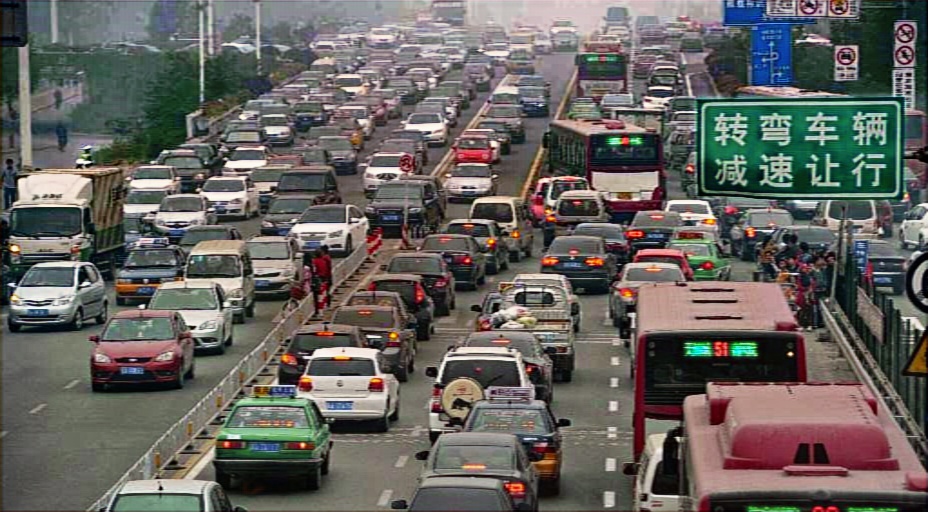}
		\\ \vskip2pt
		
		\includegraphics[width=0.19\linewidth,height=0.14\linewidth]{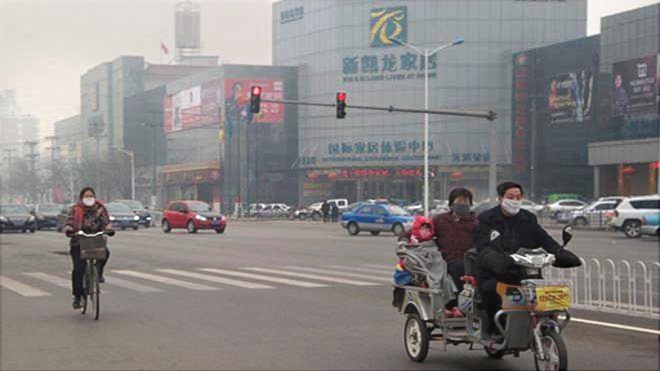}
		\includegraphics[width=0.19\linewidth,height=0.14\linewidth]{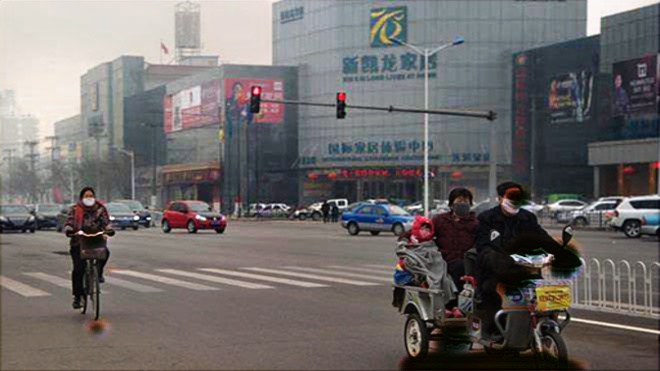}
		\includegraphics[width=0.19\linewidth,height=0.14\linewidth]{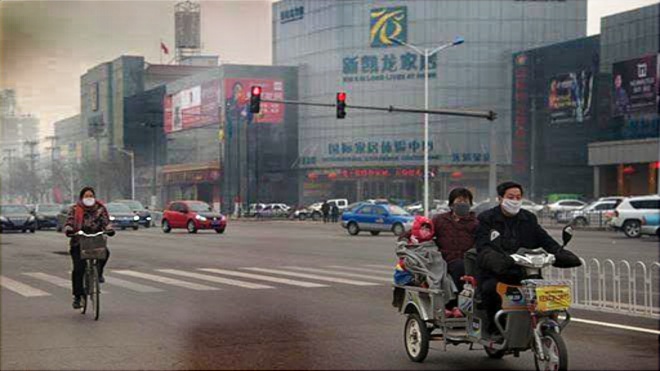}
		\includegraphics[width=0.19\linewidth,height=0.14\linewidth]{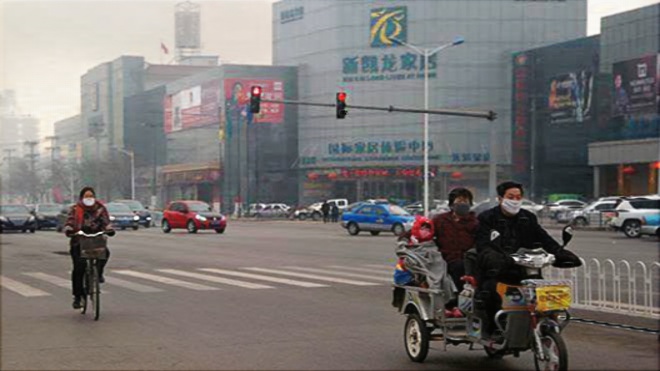}
		\includegraphics[width=0.19\linewidth,height=0.14\linewidth]{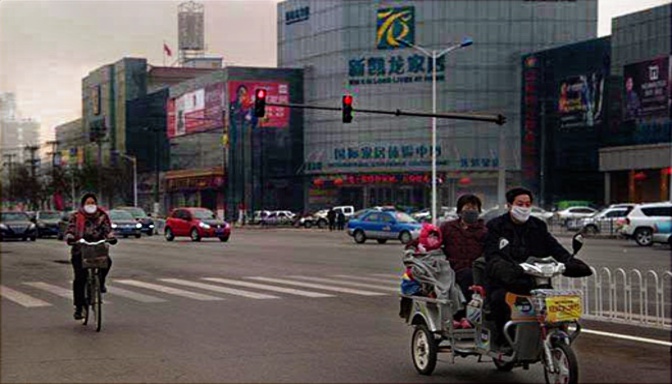}
		\\ \vskip2pt
		\includegraphics[width=0.19\linewidth,height=0.14\linewidth]{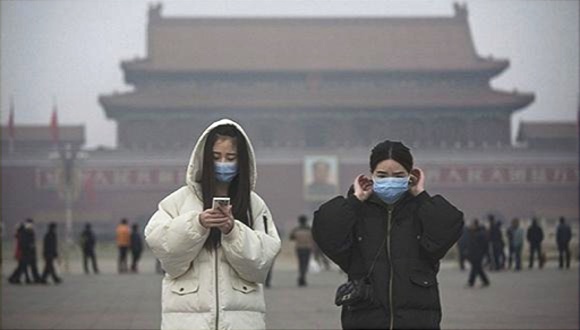}
		\includegraphics[width=0.19\linewidth,height=0.14\linewidth]{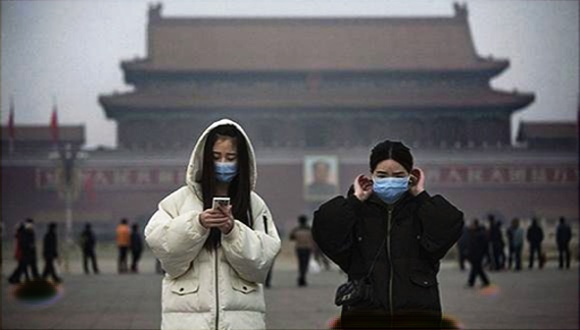}
		\includegraphics[width=0.19\linewidth,height=0.14\linewidth]{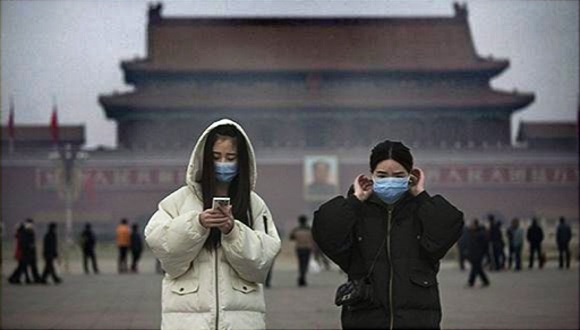}
		\includegraphics[width=0.19\linewidth,height=0.14\linewidth]{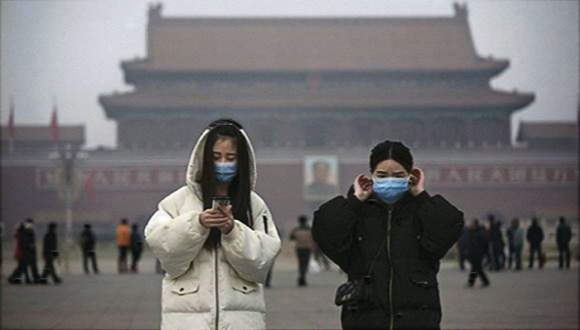}
		\includegraphics[width=0.19\linewidth,height=0.14\linewidth]{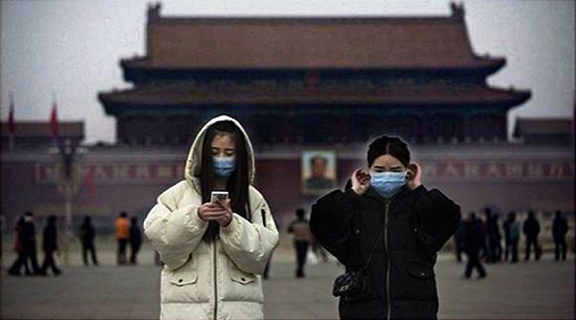}
		\\ \vskip2pt
	\end{center}
	\begin{flushleft}
		\vskip-10pt
		\hskip45pt Input\hskip60pt EPDN\cite{qu2019enhanced}\hskip45pt Grid-DeHaze\cite{liu2019griddehazenet}\hskip35pt CycGAN\cite{zhu2017unpaired}   \hskip40pt DGP-CycGAN 
	\end{flushleft}
	\vskip -20pt \caption{Sample qualitative visualizations on real hazy images of RTTS \cite{li2019benchmarking} dataset. Our results appear visually superior to fully-supervised results.}
	\label{fig:results_real_hazy}
\end{figure*}

\begin{figure*}[hp!]
	\begin{center}
	
		\includegraphics[width=0.195\linewidth]{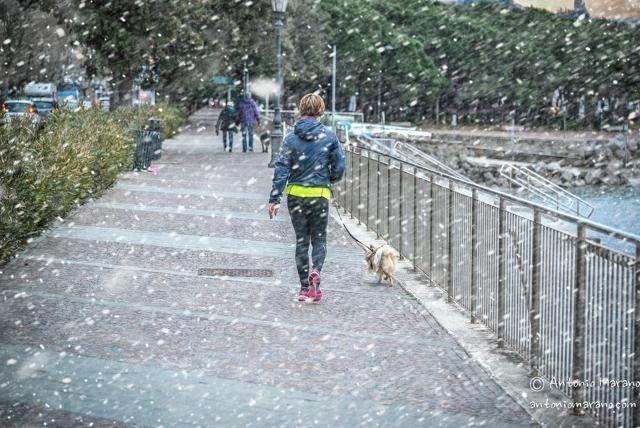}
		\includegraphics[width=0.195\linewidth]{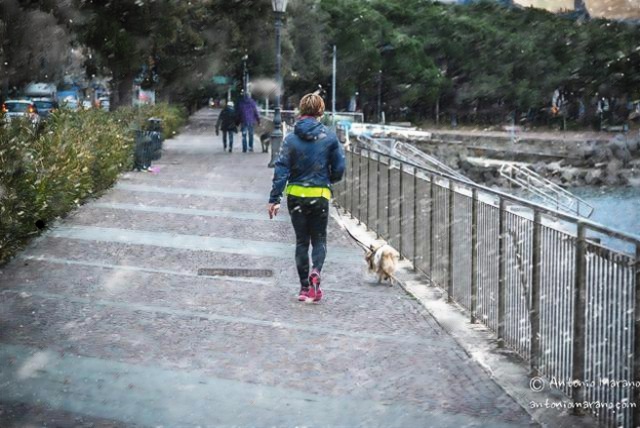}
		\includegraphics[width=0.195\linewidth]{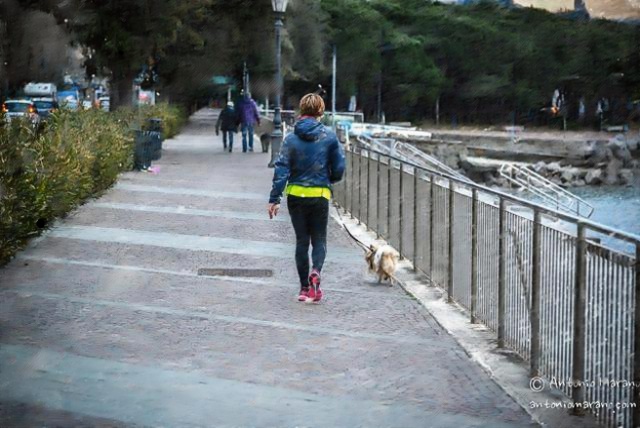}
		\includegraphics[width=0.195\linewidth]{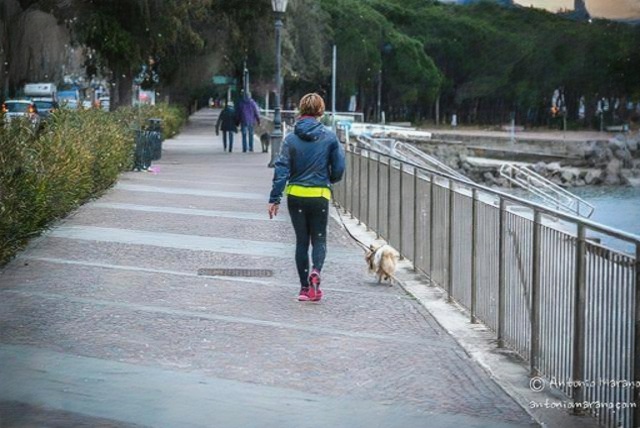}
		\includegraphics[width=0.195\linewidth]{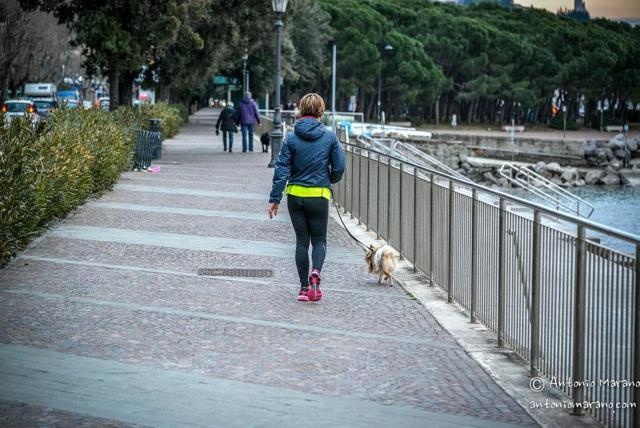}\\ \vskip2pt
		
		\includegraphics[width=0.195\linewidth]{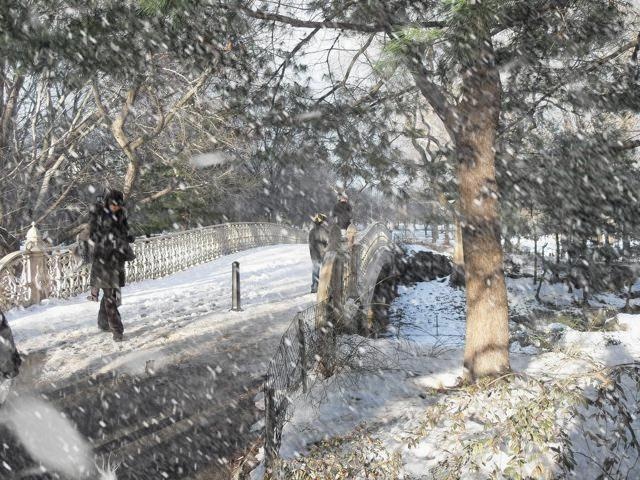}
		\includegraphics[width=0.195\linewidth]{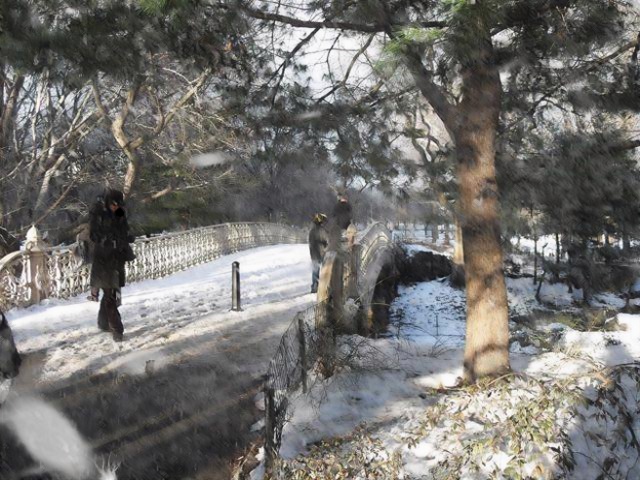}
		\includegraphics[width=0.195\linewidth]{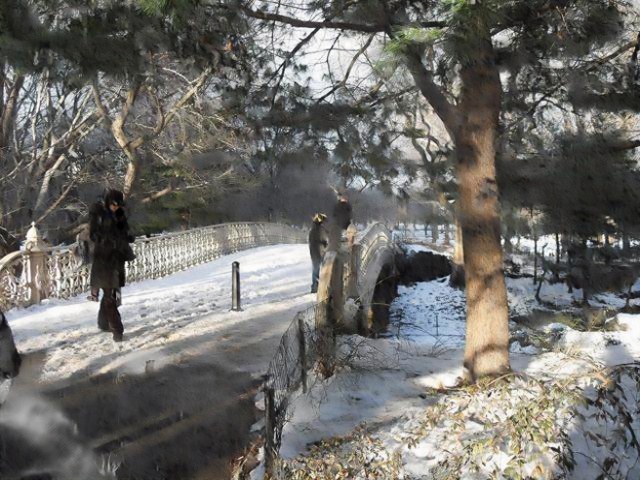}
		\includegraphics[width=0.195\linewidth]{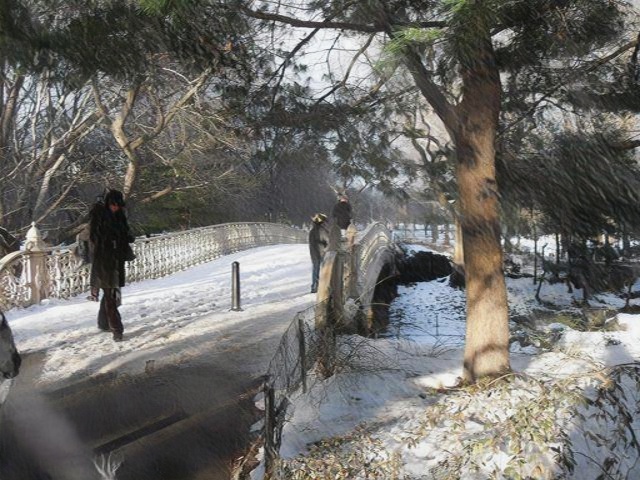}
		\includegraphics[width=0.195\linewidth]{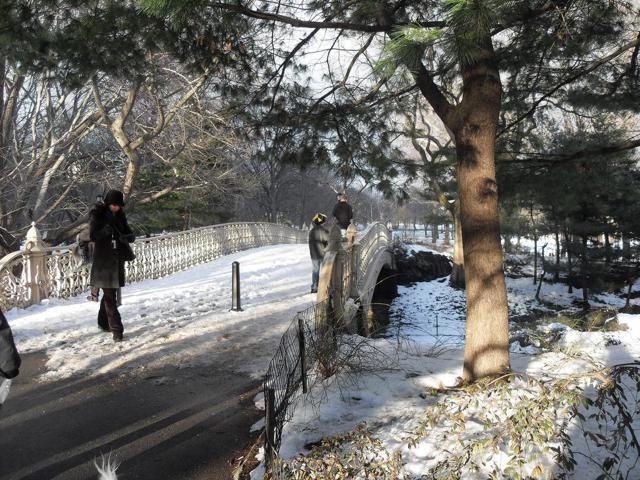}\\ \vskip2pt
	\end{center}
	\begin{flushleft}
		\vskip-10pt
		\hskip45pt Input\hskip60pt CycGAN \hskip50pt DGP-CycGAN \hskip50pt Oracle \hskip60pt Ground-truth
	\end{flushleft}
	\vskip -13pt \caption{Sample qualitative visualizations of  de-snowing on  Snow100k\cite{liu2018desnownet} dataset. Our results appear visually similar to fully-supervised results and the ground-truth.}
	\label{fig:snow_results}
\end{figure*}

\begin{figure*}[htp!]
	\begin{center}
		
		\includegraphics[width=0.195\textwidth,height=0.12\textwidth]{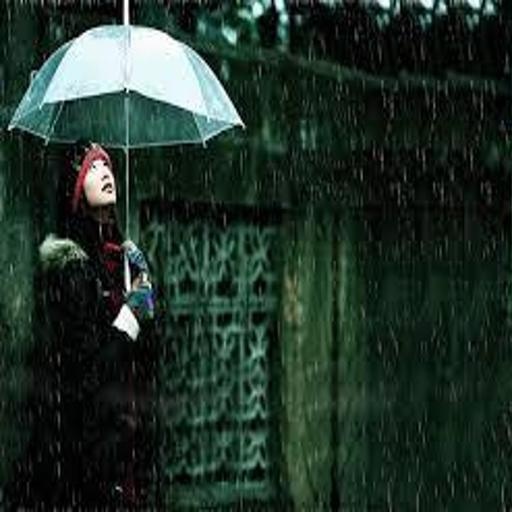}
		\includegraphics[width=0.195\textwidth,height=0.12\textwidth]{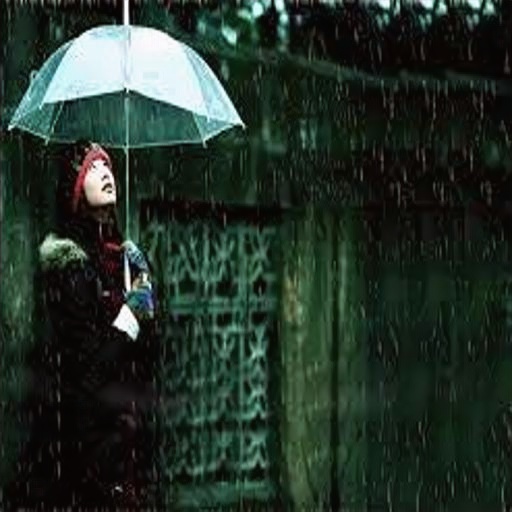}
		\includegraphics[width=0.195\textwidth,height=0.12\textwidth]{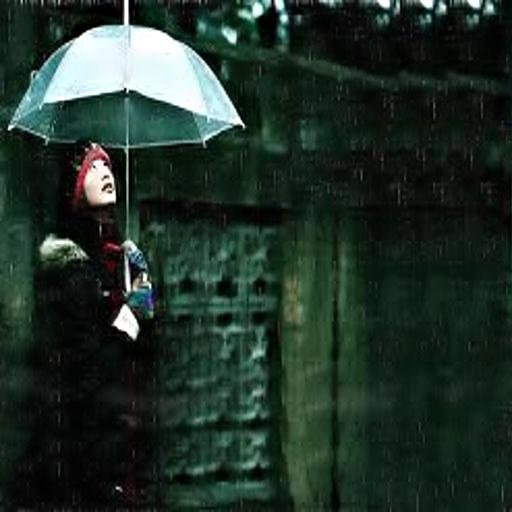}
		\includegraphics[width=0.195\textwidth,height=0.12\textwidth]{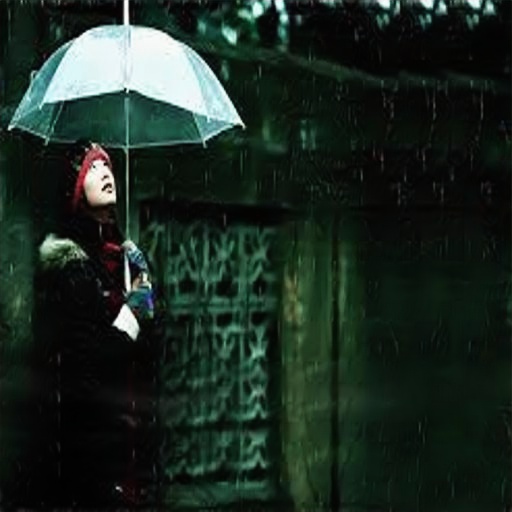}
		\includegraphics[width=0.195\textwidth,height=0.12\textwidth]{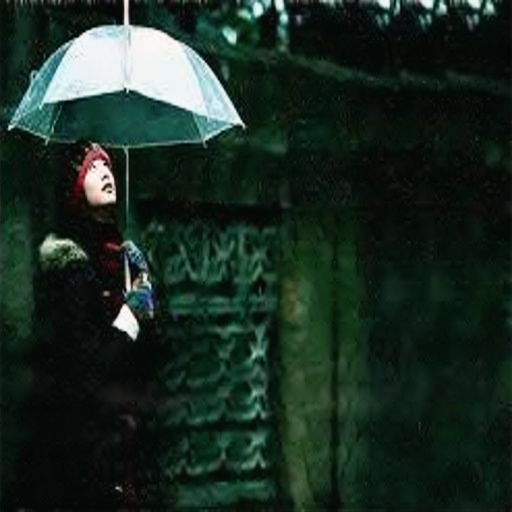}\\ \vskip3pt
		\includegraphics[width=0.195\textwidth,height=0.12\textwidth]{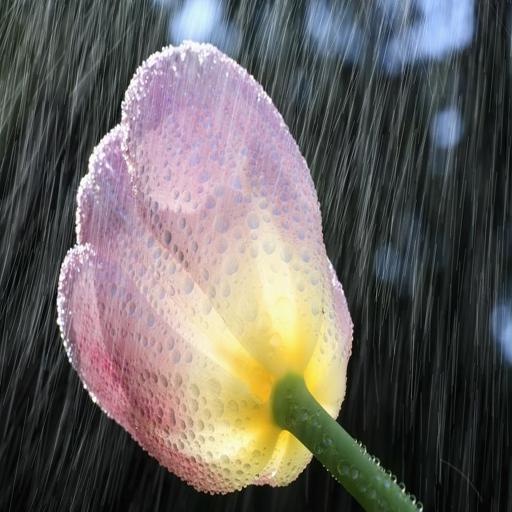}
		\includegraphics[width=0.195\textwidth,height=0.12\textwidth]{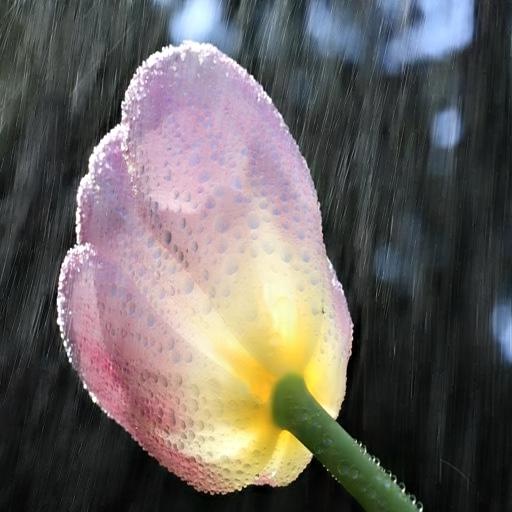}
		\includegraphics[width=0.195\textwidth,height=0.12\textwidth]{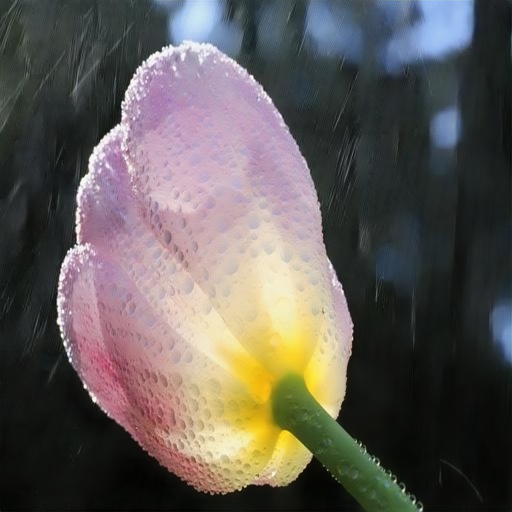}
		\includegraphics[width=0.195\textwidth,height=0.12\textwidth]{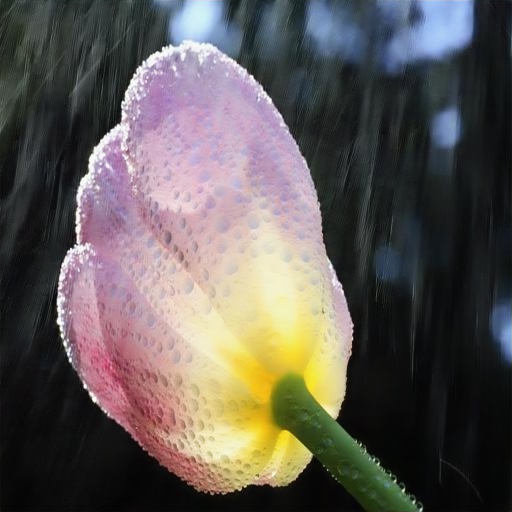}
		\includegraphics[width=0.195\textwidth,height=0.12\textwidth]{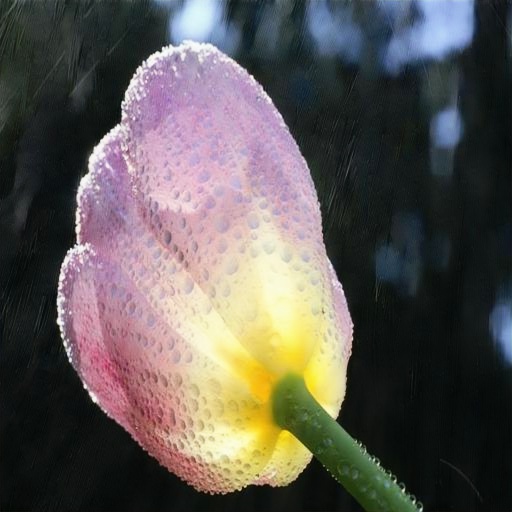}\\ 
		\begin{flushleft}
			\vskip-10pt
			\hskip45pt Input\hskip60pt CycGAN \hskip70pt PreNet \hskip50pt MSFPN \hskip60pt DGP-CycGAN
		\end{flushleft}
		\vskip-20pt
		\caption{De-raining results on sample real-world images.}
		\label{Fig:exp6}
	\end{center}
\end{figure*}

In  Figure~\ref{Fig:exp6}, we show sample results for the task of de-raining. In Figure~\ref{fig:results_real_hazy} we show sample results for the task of de-hazing on real hazy images of RTTS \cite{li2019benchmarking} dataset. Our results appear visually superior to fully-supervised method results.  In Figure~\ref{fig:snow_results} we show sample results for the task of de-snowing.


\bibliographystyle{IEEEtran}
\bibliography{egbib}
